\pdfoutput=1
\documentclass[lettersize,journal]{IEEEtran}

\usepackage{subfigure}

\usepackage{booktabs}       
\usepackage{amsfonts}       
\usepackage{nicefrac}       
\usepackage{microtype}      

\usepackage{graphicx}
\usepackage{amsmath}
\usepackage{amssymb}

\usepackage{threeparttable}
\usepackage{multirow}
\usepackage{amsmath}
\usepackage{caption}

\usepackage{algorithm}
\usepackage{algorithmic}

\usepackage{url}

\usepackage{bbding}
\usepackage{threeparttable}
\usepackage{dsfont}

\begin{document}

\title{Source-Free Unsupervised Domain Adaptation with Hypothesis Consolidation of Prediction Rationale}

\author{Yangyang Shu, Xiaofeng Cao, Qi Chen, Bowen Zhang, Ziqin Zhou, Anton van den Hengel,

~\IEEEmembership{ and Lingqiao Liu*}

\thanks{

Yangyang Shu, Qi Chen, Bowen Zhang, Ziqin Zhou, Anton van den Hengel and Lingqiao Liu are with the School of Computer Science, The University of Adelaide, Adelaide, SA 5005, Australia (e-mail: yangyang.shu@adelaide.edu.au; qi.chen04@adelaide.edu.au; b.zhang@adelaide.edu.au; ziqin.zhou@adelaide.edu.au; anton.vandenhengel@adelaide.edu.au; lingqiao.liu@adelaide.edu.au).

Xiaofeng Cao is with the School of Artificial Intelligence, Jilin University, Changchun, 130012, China,  (e-mail: xiaofeng.cao.uts@gmail.com).

* Corresponding author.
}}

\markboth{Journal of \LaTeX\ Class Files,~Vol.~14, No.~8, August~2021}%
{Shell \MakeLowercase{\textit{et al.}}: A Sample Article Using IEEEtran.cls for IEEE Journals}


\maketitle

\begin{abstract}
Source-Free Unsupervised Domain Adaptation (SFUDA) is a challenging task where a model needs to be adapted to a new domain without access to target domain labels or source domain data. The primary difficulty in this task is that the model's predictions may be inaccurate, and using these inaccurate predictions for model adaptation can lead to misleading results. To address this issue, this paper proposes a novel approach that considers multiple prediction hypotheses for each sample and investigates the rationale behind each hypothesis. By consolidating these hypothesis rationales, we identify the most likely correct hypotheses, which we then use as a pseudo-labeled set to support a semi-supervised learning procedure for model adaptation. To achieve the optimal performance, we propose a three-step adaptation process: model pre-adaptation, hypothesis consolidation, and semi-supervised learning. Extensive experimental results demonstrate that our approach achieves state-of-the-art performance in the SFUDA task and can be easily integrated into existing approaches to improve their performance. The codes are available at \url{https://github.com/GANPerf/HCPR}.

\end{abstract}


\section{Introduction}
\label{sec:intro}
The success of deep learning models in visual tasks is largely dependent on whether the training and testing data share similar distributions \cite{he2016deep, liang2020polytransform}. However, when the distribution of the testing data differs significantly from that of the training data, also known as domain shift, the performance of these models can decrease substantially \cite{tzeng2017adversarial, peng2019moment}. To mitigate the effects of domain shift and reduce the need for data annotations, Unsupervised Domain Adaptation (UDA) techniques have been developed to transfer knowledge from annotated source domains to new but related target domains without requiring annotations in the target domain \cite{hoffman2018cycada,long2018conditional,dai2020contrastively,feng2021complementary,mei2020instance}. However, most UDA-based methods rely on access to labeled source domain data during adaptation, such an access may not always be feasible due to privacy concerns. As a result,  Source-Free Unsupervised Domain Adaptation (SFUDA) \cite{liang2020we,yang2021generalized,yang2021exploiting,chen2022contrastive,yang2022attracting,zhang2022divide,karim2023c} gains much attention recently, which only requires a pre-trained model from the source domain and unlabeled data from the target domain.

The main challenge in SFUDA research is how to generate supervision solely from unlabeled data. The current approaches in SFUDA research primarily focus on either generating pseudo-labels \cite{liang2020we, yang2021generalized, yang2021exploiting, litrico2023guiding} or conducting unsupervised feature learning \cite{huang2021model, chen2022contrastive, zhang2022divide, karim2023c, litrico2023guiding} to address this issue. To generate reliable pseudo-labels, existing methods \cite{liang2020we, yang2021generalized, yang2021exploiting} often utilize the distribution of the target domain data to refine the initial predictions from the source domain, i.e., via clustering \cite{liang2020we} or using the predictions of neighboring samples \cite{yang2021exploiting, litrico2023guiding}. On the other hand, unsupervised feature learning, such as contrastive learning, is often employed as an auxiliary task to encourage the features to adapt to the target domain \cite{huang2021model, chen2022contrastive, zhang2022divide, karim2023c, litrico2023guiding}. 

In our study, we propose a novel approach to tackle the challenge of SFUDA. Our strategy involves deferring the utilization of label predictions to update the model in the early stages and carefully selecting the most reliable predictions to construct a pseudo-labeled set. The key innovation of our approach lies in considering multiple prediction hypotheses for each sample, accommodating the possibility of multiple potential labels for each data point. We treat each label assignment as a hypothesis and delve into the rationale and supporting evidence behind each prediction. We utilize a representation derived from GradCAM~\cite{selvaraju2017grad} to encode the rationale for predicting an instance to a hypothetical label. Our methodology is inspired by the belief that assessing the correctness of a prediction can be more reliable by analyzing the reasoning behind a particular prediction, rather than solely relying on prediction probabilities.
Subsequently, we develop a consolidation method to determine the most trustworthy hypothesis and utilize it as the labeled dataset in a semi-supervised learning framework. By employing this technique, we effectively transform the SFUDA problem into a conventional semi-supervised learning problem.

Concretely, our approach consists of three key steps: model pre-adaptation, hypothesis consolidation, and semi-supervised learning. We have empirically observed that pre-adapting the model can enhance the effectiveness of the second step. To accomplish this, we introduce a straightforward objective that encourages prediction smoothness from the network. In the final step, we leverage the widely-used FixMatch~\cite{sohn2020fixmatch} algorithm as our chosen semi-supervised learning method. Through extensive experimentation, we demonstrated the clear advantages of our approach over existing methods in the SFUDA domain and show that the proposed method can be easily integrated into existing approaches to bring improvement.

\section{Related Work}
\label{related_work}
\subsection{UDA} 
Unsupervised domain adaptation aims to transfer knowledge learned from a labeled source domain to an unlabeled target domain. Various approaches have been proposed to address this task, including discrepancy minimization~\cite{tzeng2014deep, ganin2015unsupervised, long2015learning}, adversarial learning~\cite{hoffman2018cycada,long2018conditional,tzeng2017adversarial,vu2019advent}, and contrastive learning~\cite{dai2020contrastively,kang2019contrastive}. Recently, self-training using labeled source data and pseudo-labeled target data has emerged as a prominent approach in unsupervised domain adaptation (UDA) research~\cite{feng2021complementary,mei2020instance,xie2020self,yu2021dast,zou2018unsupervised}. However, these methods typically rely on access to the source data, making them inapplicable when source data is unavailable.

\subsection{SFUDA.} Source-free unsupervised domain adaptation involves adapting a pre-trained model from a source domain to a target domain without access to source data+labels or target labels. Existing SFUDA methods can be broadly categorized into two classes: i) Label Refinement: Methods such as SHOT~\cite{liang2020we}, G-SFDA~\cite{yang2021generalized}, NRC~\cite{yang2021exploiting}, and GPL~\cite{litrico2023guiding} focus on refining pseudo labels. SHOT generates pseudo labels using centroids obtained in an unsupervised manner. G-SFDA, NRC, and GPL refine pseudo labels through consistent predictions and nearest neighbor knowledge aggregation from local neighboring samples. ii) Contrastive Feature Learning: Approaches like HCL~\cite{huang2021model}, C-SFDA~\cite{karim2023c}, AdaContrast~\cite{chen2022contrastive}, GPL~\cite{litrico2023guiding}, and DaC~\cite{zhang2022divide}. HCL and C-SFDA use a contrastive loss similar to moco~\cite{he2016deep}, where positive pairs consist of augmented query samples and negatives are other samples. AdaContrast and GPL exclude same-class negative pairs based on pseudo labels. DaC divides the target data into source-like and target-specific samples, computes source-like class centroids, and generates negative pairs using these centroids. These methods aim to tackle SFUDA by refining pseudo labels or leveraging contrastive feature learning, demonstrating the potential of different strategies in addressing the challenges of adapting models without access to labeled source data or target label.

\section{Method}
\begin{figure}[htbp]
	\centering
	\includegraphics[width=3.5in]
 {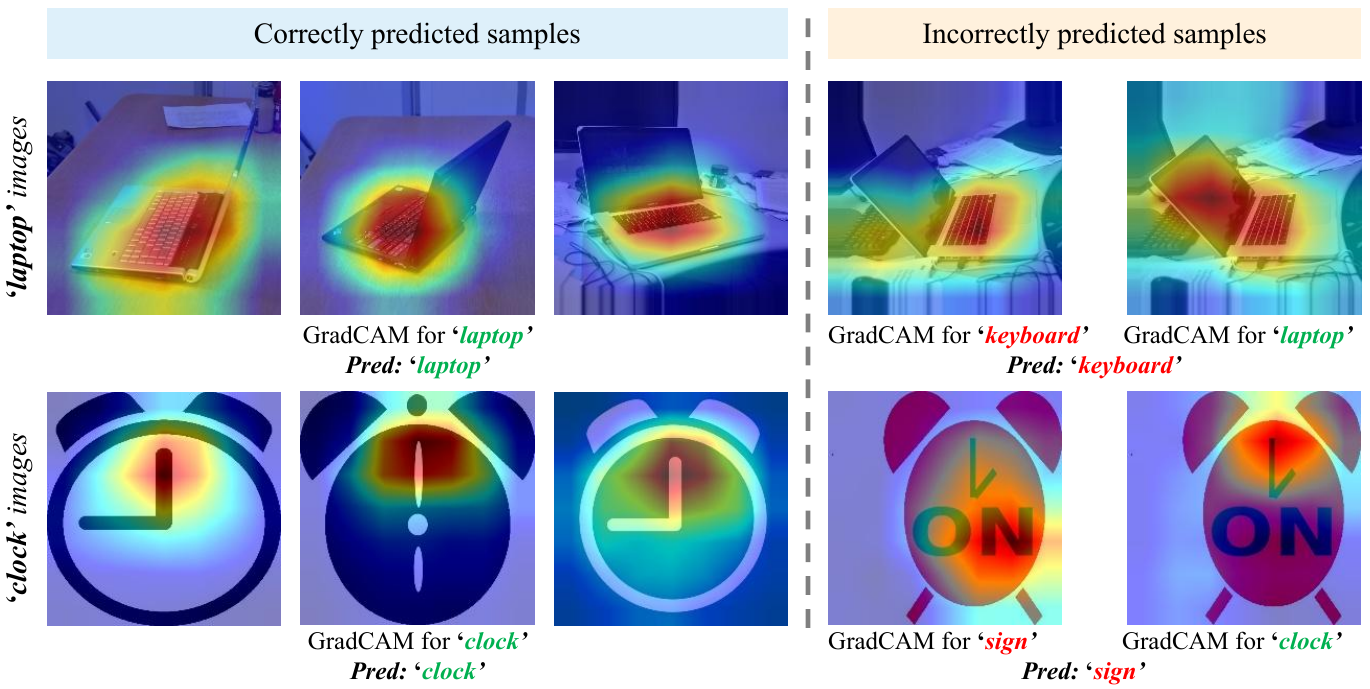}
	\caption{ The visualizations illustrate the GradCAM~\cite{selvaraju2017grad} for predicting the image to a specific class. In the right-half section, it can be observed that even though the prediction is incorrect, the obtained rationale (region highlighted in the GradCAM) based on the correct label remains reasonable and resembles the rationale of the corresponding class depicted in the left-half section.}
	\label{visual_rationale}
\end{figure}
In the source-free unsupervised domain adaptation (SFUDA) setting, only pretrained source models and unlabeled data in the target domain are given. The task is to adapt the model to the target domain by using unlabeled target data only. 
Our approach sequentially applies three steps as described in Sec.~\ref{method_PPLG}), Sec.~\ref{method_rationale}) and Sec.~\ref{method_fixmatch}). 


\begin{figure*}[tbp]
	\centering
	\includegraphics[width=5.5in]{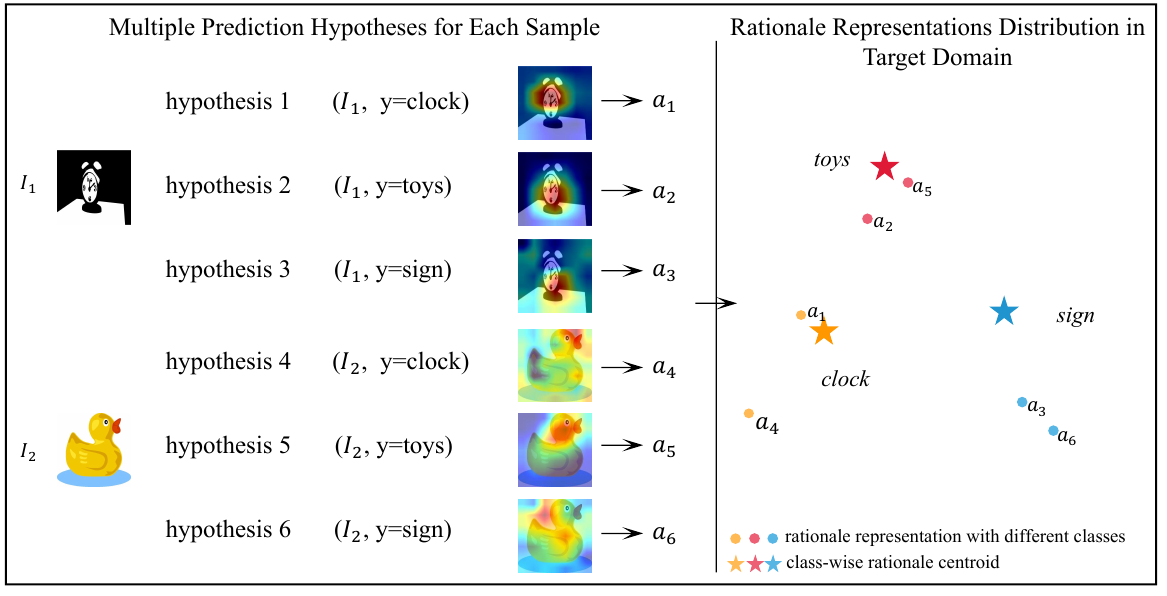}
 \caption{In our method, we generate multiple prediction hypotheses based on the posterior probability of the current model. An image $I$ and its hypothetical label form a hypothesis, for example, ($I$, $y=\text{clock}$). For each hypothesis, GradCAM is calculated based on the hypothetical label, resulting in the corresponding rationale representation $a$. Subsequently, we calculate the centroid for the rationale representation of each class.}
	\label{framework_rationale}
\end{figure*}

\subsection{Model Pre-adaptation via Encouraging Smooth Prediction}
\label{method_PPLG}
The first step of our approach is to make an initial adaptation to reduce the domain gap. We empirically find such a step can be beneficial for the following steps. We develop a pre-adaptation strategy by encouraging a smooth prediction on the data manifold. \footnote{Other pre-adaptation approaches may also work, such as the method in \cite{liang2020we}, please refer to Sec.~\ref{exp: SHOT+ours} for more experimental evidence.} Specifically, we create a memory $Q\in \mathbb{R}^{N_q \times d}$ to store $N_q$ randomly sampled image embedding and update it after each batch training. Then for each target sample $x_i$, we find the $z$-nearest neighbor $\mathcal{NN}(x_i)$ and $z$-samples $\mathcal{FN}(x_i)$ that are furthest to $x_i$ based on the Euclidean distance between the image embedding of $x_i$ and embedding in $Q$ ($d=256$ and $z=3$ in our implementation). Then we optimize the following objective:
\begin{align}\label{eq: PLG}
    \mathcal{L}_{PA} &= \mathcal{L}_{SM} + \lambda\mathcal{L}_{FAR} = \sum_{i=1}^{N_B}\sum_{x'_j \in \mathcal{NN}(x_i)}KL(p(x_i), p(x'_j)) \nonumber\\& + \lambda \sum_{i=1}^{N_B}\sum_{x'_j \in \mathcal{FN}(x_i)} p(x_i)^\top p(x'_j) , 
\end{align} where KL represents Kullback-Leibler divergence and $p$ denotes the posterior probability predicted from the source model. $N_B$ is the number of samples within a mini-batch. The first term is used to ensure similar samples have similar predictions. However, using the first term alone may lead to a trivial solution that assigns identical prediction for every instance. Thus we use the second term to counter-act it as it ensures that the least similar samples should have divergent posterior probabilities, i.e., the inner product between posterior should close to zero.

\subsection{Hypothesis Consolidation from Prediction Rationale}
\label{method_rationale}
After pre-adaptation, the model generally exhibits improved adaptation to the target domain. However, there may still be instances where the model produces incorrect predictions, making it challenging to rectify misclassifications solely based on predicted posterior probabilities. 
Therefore, in the second step, we explore a more robust methodology for analyzing predictions.

We begin by considering multiple prediction hypotheses for each individual instance. Specifically, for each instance, we consider the top $\tilde{k}$ classes with the highest posterior probabilities as potential prediction hypotheses, denoted as $(x_i,y^h_{ik})$, $k\in$ top $\tilde{k}$. In other words, we acknowledge the correct class label could exist within one of these top $\tilde{k}$ classes, even though we do not know which one.

To further analyze each hypothesis $(x_i,y^h_{ik})$, we calculate the GradCAM \cite{selvaraju2017grad} to identify the regions that contribute to supporting the prediction for $y^h_{ik}$, resulting in a representation called the rationale representation $a_{ik}$. This rationale representation encodes the evidence supporting the corresponding hypothesis. Drawing inspiration from prior work \cite{shu2022improving,shu2023learning}, we formally calculate $a_{ik}$ using following equation:  

\begin{footnotesize}
    \begin{align}\label{eq: ratioanle} 
    a_{ik} = \frac{1}{HW}\sum_{m=1}^H\sum_{n=1}^W\left(\left[\frac{\partial logit(y^h_{ik})}{\partial [\phi(x_i)]_{m,n}}^\top[\phi(x_i)]_{m,n}\right]_+\cdot[\phi(x_i)]_{m,n}\right), 
\end{align}
\end{footnotesize}where $a_{ik}\in\mathbb{R}^{d'}$, $\phi(x_i)\in \mathbb{R}^{H \times W \times d'}$ is the feature map of the last convolutional layer of the network with $H$ height, $W$ width, and $d'$ channels. $[\phi(x_i)]_{m,n}\in \mathbb{R}^{d'}$ is the feature vector located at the $(m,n)$-th grid. $logit(y^h_{ik})$ is the logit for class $y^h_{ik}$, $[\cdot]_+=max(\cdot, 0)$. $\left[\frac{\partial logit(y^h_{ik})}{\partial [\phi(x_i)]_{m,n}}^\top[\phi(x_i)]_{m,n}\right]_+$ is equivalent to GradCAM value at the $(m,n)$-th grid. Essentially, the calculation of $a_{ik}$ performs weighted average pooling over $\phi(x_i)$ according to the GradCAM. Figure~\ref{visual_rationale} shows the GradCAM calculated from different hypotheses for the same image. Upon observation, we notice that even if the ground-truth class is not ranked as the top prediction by the model, its associated rationale remains reasonable and similar to the common rationale patterns for the corresponding class. This inspires us to leverage this observation to analyze the model's current predictions. For example, if an instance has a prediction hypothesis that exhibits a rationale similar to the corresponding class's common rationale but is not ranked as the top prediction, then the top prediction may not be correct.

Formally, we calculate the class-wise rationale centroid as the average rationale representation from each hypothetical class, representing the common rationale for each class: 
\begin{align}
    \bar{a}_c = \frac{\sum_{ik}\mathds{1}(y^h_{ik}=c) a_{ik}}{\sum_{ik}\mathds{1}(y^h_{ik}=c)}, 
\end{align} where $c$ represents a class and $\mathds{1}(y^h_{ik}=c) = 1$ if $k=c$. The idea of using multiple hypotheses with the rationale representation is illustrated in Figure~\ref{framework_rationale}.

Next, we generate a ranking index $r_{ik}$ for each prediction hypothesis $(x_i,a_{ik},y^h_{ik})$ by ranking the Euclidean distance between $a_{ik}$ and its corresponding rationale centroid $\bar{a}_{y^h_{ik}}$, i.e., the centroid for class $y^h_{ik}$, in the ascending order. For each instance $x_i$, we obtain $\tilde{k}$ ranking indices ${r_{ik}}$, $k\in$ top $\tilde{k}$ classes, one for each hypothesis.
Then, a hypothesis $\{x_i, y_{ik'}\}$ is considered reliable if it satisfies the following two conditions:
(1) $r_{ik'} < \tau_1$, indicating the rationale for $\{x_i, y_{ik'}\}$ is typical as its rationale representation is close to the rationale centroid.
(2) $r_{ij} > \tau_2 ~~ \forall j \neq k'$, where $\tau_2 > \tau_1$ are two predefined ranking thresholds. The second condition ensures that there are no conflicting hypotheses, i.e., no other hypothesis is likely to be true for the same instance as their rationale appears to be unusual.

\begin{figure*}[tbp]
	\centering
	\includegraphics[width=5.0in]{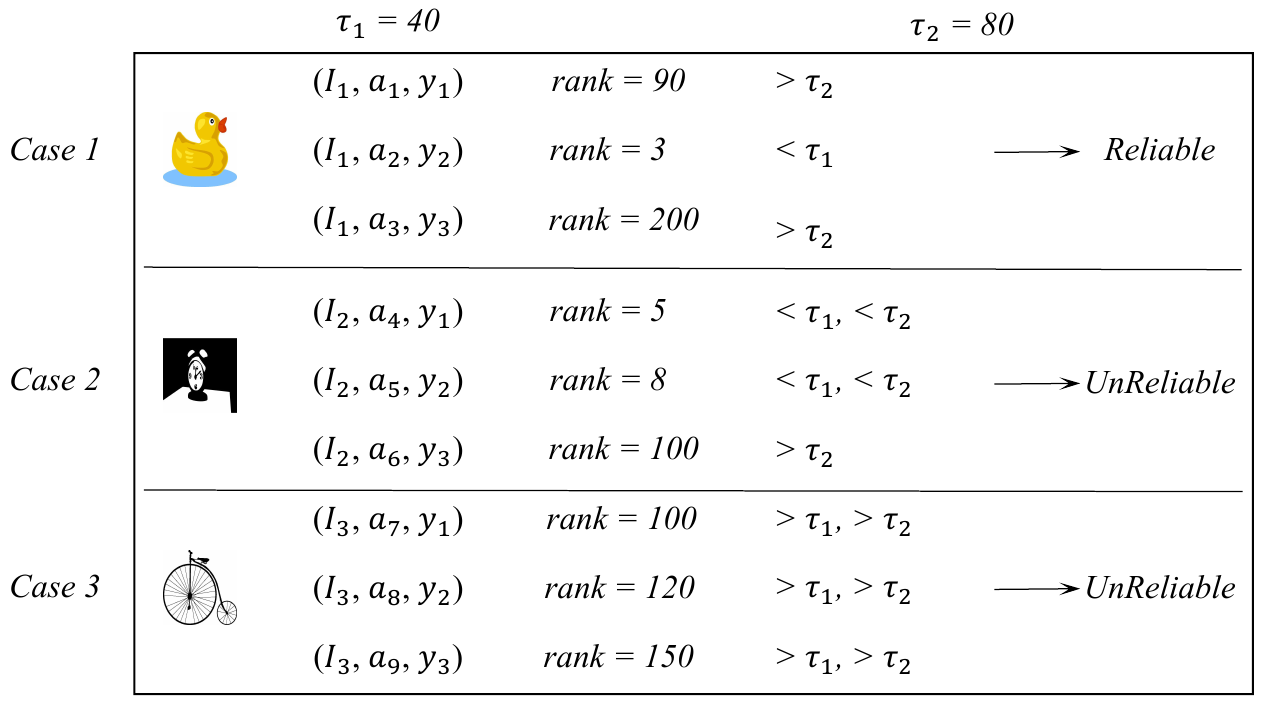}
	\caption{These examples demonstrate the generation of reliable hypotheses. In Case 1, the rank ID of the second hypothesis derived from the image is lower than $\tau_1$, while all other hypotheses from the same image have ranks larger than $\tau_2$. Consequently, the second hypothesis of $I_1$ is selected as a reliable hypothesis. In Case 2, no hypothesis is selected because it has two hypotheses with rank IDs less than $\tau_2$, indicating a conflict between those hypotheses. Similarly, Case 3 is not selected because none of its hypotheses has rank IDs lower than $\tau_1$.}
	\label{framework_rank}
 \end{figure*}
With those criteria, we can collect a set of reliable hypotheses $\mathcal{P}$ as samples with their corresponding hypothetical labels. 
Representative examples of this procedure are depicted in Figure~\ref{framework_rank}. It is important to note that in the second step, we aim to select the most reliable hypothesis rather than correcting hypotheses. This is because we believe that the task of correcting predictions or hypotheses can be better accomplished through the use of semi-supervised learning, which allows for the gradual propagation of pseudo-labels.

By focusing on identifying the most reliable hypothesis based on the proximity of the rationale representation to the rationale centroid and the absence of conflicting rationales, we can create a high-quality set of pseudo-labeled samples (see Section~\ref{app_qq}). These pseudo-labels can then be used in a semi-supervised learning framework to refine the model's predictions and gradually improve its performance.

\subsection{Semi-Supervised Learning }
\label{method_fixmatch}
After completing the second step of hypothesis consolidation, we obtain a reliable pseudo-label set $\mathcal{P}$, while the remaining samples are treated as the unlabeled set $\mathcal{U}$. At this stage, we are ready to apply a semi-supervised algorithm to perform the final step of adaptation. For this purpose, we utilize one of the state-of-the-art semi-supervised methods, FixMatch~\cite{sohn2020fixmatch}, which combines consistency regularization and pseudo-labeling to address this task.

Specifically, we start by sampling a labeled mini-batch $\mathcal{B}_l$ from the reliable pseudo-label set $\mathcal{P}$ and an unlabeled batch $\mathcal{B}_u$ from the unlabeled set $\mathcal{U}$. We then optimize the following objective function using these batches:
\begin{align}\label{eq: fixmatch}
    \mathcal{L}_{FM} &= \sum_{x_b\in \mathcal{B}_l}CE(\hat{y}_b, p(\mathcal{A}_w(x_b)))\nonumber\\& + \sum_{x_u\in \mathcal{B}_u}  \mathds{1}(\max(p(\mathcal{A}_w(x_u)))\geq\tau) CE(\hat{y}_u, p(\mathcal{A}_s(x_u))) , 
\end{align} where $\hat{y}_u=\mathop{\arg\max}\limits_c p(y=c|\mathcal{A}_w(x_u))$. $\mathcal{A}_w(\cdot)$ and $\mathcal{A}_s(\cdot)$ are the weakly-augmented and strongly-augmented operations, respectively. $\tau$ is the threshold defined in FixMatch to identify reliable pseudo-label (we set the same with FixMatch as 0.95), and $CE$ is the cross-entropy between two probability distributions. 

We present the overall training process of our proposed SFUDA method in Algorithm~\ref{alg_l1}.
\begin{algorithm}[htbp]
	\caption{SFUDA with
Hypothesis Consolidation of Prediction Rationale}
	\label{alg_l1}
	\begin{algorithmic}
		\REQUIRE Unlabeled target data $D_t$, pre-trained model, memory $Q$ stores $N_q$ randomly sampled image embedding, two ranking thresholds $\tau_1$ and $\tau_2$; the number of steps of updating model pre-adaptation $K_1$, the number of steps of updating semi-supervised learning $K_2$.
		
			\FOR {$K_1$ steps}
			{
				\STATE Sample a mini-batch of $N_B$ training data $\{x_i\}_{i=1}^{N_B}$ from $D_t$, find $z$-nearest neighbor $\mathcal{NN}(x_i)$ and $z$-furthest neighbor $\mathcal{FN}(x_i)$.
				\STATE Update model by Eq. 1.
                    \STATE Update memory $Q$
			}
			\ENDFOR
			\STATE For each sample $x_i$ in $D_t$, calculate rationale representation of each class via Eq. 2.
                \STATE Calculate class-wise rationale centroid for each class via Eq. 3.
                \STATE Collect a set of reliable hypotheses with their corresponding hypothetical labels as a reliable pseudo-label set $\mathcal{P}$, and the remaining samples as the unlabeled set $\mathcal{U}$.
			\FOR {$K_2$ steps}
			{
			    
			    \STATE Sample a labeled mini-batch $\mathcal{B}_l$ from $\mathcal{P}$ and an unlabeled batch $\mathcal{B}_u$ from $\mathcal{U}$.
				\STATE Update model by Eq. 4. 
				
			}
			\ENDFOR
	
	\end{algorithmic}
\end{algorithm}

\section{Experiments}
\subsection{Datasets}
\textbf{Office-Home}~\cite{venkateswara2017deep} consists of 15,500 images categorized into 65 classes. It includes four distinct domains: Real-world (Rw), Clipart (Cl), Art (Ar), and Product (Pr). To evaluate the proposed method, researchers perform 12 transfer tasks on this dataset, involving adapting models across the four domains. The evaluation reports each domain shift Top-1 and the average Top-1 accuracy. Originally, the \textbf{DomainNet} dataset~\cite{peng2019moment} consisted of over 500,000 images, including six domains and 345 classes. For our evaluation, we follow the approach described in \cite{saito2019semi} and focus on four domains: Real World (Rw), Sketch (Sk), Clipart (Cl), and Painting (Pt). We assess our proposed method on seven domain shifts within these four domains. \textbf{VisDA-C}~\cite{peng2017visda} contains 152,000 synthetic images from the source domain and 55,000 real object images from the target domain. It consists of 12 object classes, and there is a significant synthetic-to-real domain gap between the two domains. Our evaluation reports per-class Top-1 accuracies, as well as the average Top-1 accuracy on this dataset.

\subsection{Implementation Details}
\label{app_implementation}
To ensure fair comparisons with previous work~\cite{liang2020we, chen2022contrastive, karim2023c}, we employ the ResNet-50~\cite{he2016deep} as the network backbone for the Office-Home and DomainNet datasets, and ResNet-101 for the VisDA-C dataset. The network architecture follows the same configuration as SHOT~\cite{liang2020we}. Specifically, we replace the original fully connected (FC) layer in ResNet-50/101 with a bottleneck layer of 256 dimensions and apply batch normalization~\cite{ioffe2015batch}. This modified setup serves as the feature extractor+projector head, producing feature representations and embedding of dimensions $d^{\prime}=2048$ and $d=256$, respectively. Additionally, we include an extra fully connected layer with weight normalization~\cite{salimans2016weight} as a task-specific classifier. 

In the first step of model pre-adaptation, we use a batch size of 64. The value of $\lambda$ is set as $\lambda=\lambda_0\cdot(1+10\cdot p^{\prime})^{-5}$, where $\lambda_0=1$, and $p^{\prime}$ represents the training progress variable ranging from 0 to 1, calculated as $\frac{iter}{max\_iter}$. In the second step of hypothesis consolidation, we set the number of nearest/furthest neighbor per instance $z$ as 3, and set hypothesis per instance $\tilde{k}$ as 4, respectively. The ranking thresholds $\tau_1$ and $\tau_2$ are determined as a percentage of the total number of samples on the three datasets, specifically set at 0.8\% and 1.6\%. In the third step of semi-supervised learning, we set the size of $\mathcal{B}_l$ and $\mathcal{B}_u$ to 64. 

We use the SGD optimizer with a momentum of 0.9 and a weight decay of $1e^{-3}$ for all datasets. The learning rate is set as $1e^{-4}$ for all datasets, except for the bottleneck layer and the additional fully connected layer, where it is set as $1e^{-3}$. We train for 40 epochs on the Office-Home and DomainNet datasets, where 9 epochs are dedicated to the model pre-adaptation. For the VisDA-C dataset, we train for 15 epochs, with 7 epochs allocated for the model pre-adaptation. All images from the datasets undergo augmentation, including weak and strong augmentation. Weak augmentation involves a standard flip-and-shift augmentation strategy, while strong augmentation is similar to the approach used in the work of \cite{sohn2020fixmatch}.


\begin{table}[htbp]
\caption{Ablation study of the proposed components calculated by average accuracy (\%) on the \textbf{Office-Home} (O-H), \textbf{DomainNet} (DN) and \textbf{VisDA-C} datasets. PA stands for model pre-adaptation (Sec.~\ref{method_PPLG}), HCPR (Sec.~\ref{method_rationale}) stands for hypothesis consolidation from prediction rationale, FM stands for FixMatch (Sec.~\ref{method_fixmatch} ).}
\resizebox{\linewidth}{!}{
\label{ablation_component}
\begin{tabular}{lcccccc}
\toprule
\#&PA & HCPR & FM & O-H & DN & VisDA-C \\ \midrule
0&$\times$     & $\times$        & $\times$        & 60.2        & 55.6      & 46.6 \\
1&$\times$     & $\times$        & $\checkmark$        & 64.2        & 60.6      & 62.3 \\
2&$\times$     & $\checkmark$         & $\checkmark$        & 68.6        & 70.6      & 85.2    \\ \midrule
3&$\checkmark$   & $\times$           &$\times$           & 72.1        & 67.4      & 86.2    \\
4&$\checkmark$   & $\checkmark$         &$\times$           & 72.7        & 69.6      & 87.5    \\ \midrule
5&$\checkmark$   & $\times$          & $\checkmark$        & 72.2        & 67.5      & 86.2    \\
6&$\checkmark$   & $\checkmark$         & $\checkmark$        & \textbf{73.6}        & \textbf{72.5}      & \textbf{88.6}    \\\bottomrule
\end{tabular}}
\end{table}

\begin{table*}[tp]
\caption{Accuracy (\%) on medium-sized \textbf{Office-Home} dataset (ResNet-50). ``SF'' denotes source-free. We highlight the best results.}
\resizebox{\linewidth}{!}{
\label{exp_office-home}
\begin{tabular}{l|c|cccccccccccc|c}
\toprule
Method                   & SF & Ar$\rightarrow$Cl                        & Ar$\rightarrow$Pr                        & Ar$\rightarrow$Rw                        & Cl$\rightarrow$Ar                        & Cl$\rightarrow$Pr                        & Cl$\rightarrow$Rw                        & Pr$\rightarrow$Ar                        & Pr$\rightarrow$Cl                        & Pr$\rightarrow$Rw                        & Rw$\rightarrow$Ar                        & Rw$\rightarrow$Cl                        & Rw$\rightarrow$Pr                        & Avg.                        \\ \midrule
ResNet-50~\cite{he2016deep}                & $\times$  & 34.9                        & 50.0                        & 58.0                        & 37.4                        & 41.9                        & 46.2                        & 38.5                        & 31.2                        & 60.4                        & 53.9                        & 41.2                        & 59.9                        & 46.1                        \\
GSDA~\cite{hu2020unsupervised}                & $\times$  & 61.3                       & 76.1                      & 79.4                        & 65.4                        & 73.3                       &  74.3                     &   65.0                     &53.0                         &   80.0                     & 72.2                       &  60.6                      & 83.1                        &  70.3                      \\
RSDA~\cite{gu2020spherical}                & $\times$  & 53.3                       & 77.7                      &   81.3                      &  66.4                       &  74.0                      & 76.5                      &   67.9                     & 53.0                        &   82.0                     & 75.8                       &   57.8                     &  85.4                       &70.9                        \\
SRDC~\cite{tang2020unsupervised}                & $\times$                         &  52.3                     &  76.3                       &   81.0                      & 69.5                       &76.2                       & 78.0                       &\textbf{68.7}                         &  53.8                      & 81.7                       &  76.3                      &  57.1                       &  85.0      &71.3                \\
FixBi~\cite{na2021fixbi}                & $\times$  &58.1                        & 77.3                      &  80.4                       &  67.7                       &   79.5                     & 78.1                      &  65.8                      & 57.9                        &   81.7                     & 76.4                       & 62.9                       & \textbf{86.7}                        &72.7                        \\\midrule
G-SFDA~\cite{yang2021generalized}                   & $\checkmark$  & 57.9                        & 78.6                        & 81.0                        & 66.7                        & 77.2                        & 77.2                        & 65.6                        & 56.0                        & 82.2                        & 72.0                        & 57.8                        & 83.4                        & 71.3                        \\
SHOT~\cite{liang2020we}                     & $\checkmark$  & 56.9                        & 78.1                        & 81.0                        & 67.9                        & 78.4                        & 78.1                        & 67.0                        & 54.6                        & 81.8                        & 73.4                        & 58.1                        & 84.5                        & 71.6                        \\
SHOT++~\cite{liang2021source} & $\checkmark$  & 57.9                        & 79.7                        & 82.5                        & 68.5                        & 79.6                        & 79.3                        & 68.5                        & 57.0                        & 83.0                        & 73.7                        & 60.7                        & 84.9                        & 73.0                        \\
NRC~\cite{yang2021exploiting}                      & $\checkmark$  & 57.7                        & \textbf{80.3} & 82.0                        & 68.1                        & 79.8                        & 78.6                        & 65.3                        & 56.4                        & 83.0                        & 71.0                        & 58.6                        &  85.6 & 72.2                        \\
CoWA~\cite{lee2022confidence}                & $\checkmark$  & 56.9                        & 78.4                        & 81.0                        & 69.1                        &  80.0 & 79.9 & 67.7 & 57.2                        & 82.4                        & 72.8                        & 60.5                        & 84.5                        & 72.5                        \\
HCL~\cite{huang2021model}                      & $\checkmark$ & \textbf{64.0}                        & 78.6                        & 82.4 & 64.5                        & 73.1                        &  80.1 & 64.8                        & \textbf{59.8} & 75.3                        &  \textbf{78.1} & \textbf{69.3} &  81.5 & 72.6                        \\
AaD~\cite{yang2022attracting} & $\checkmark$ & 59.3                        & 79.3                        & 82.1                        & 68.9                        & 79.8                       & 79.5                        & 67.2                        & 57.4                        & 83.1                        & 72.1                        & 58.5                        & 85.4                        & 72.7                        \\
DaC~\cite{zhang2022divide}                      & $\checkmark$ & 59.1                        & 79.5                        & 81.2                        & 69.3 & 78.9                        & 79.2                        & 67.4                        & 56.4                        & 82.4                        & 74.0 & 61.4                        & 84.4                        & 72.8 \\
VMP~\cite{jing2022variational} & $\checkmark$ & 57.9                        & 77.6                        &  82.5                       & 68.6 & 79.4                        & \textbf{80.6}                       & 68.4                        &   55.6                      & 83.1                        &75.2  & 59.6                        & 84.7                        & 72.8 \\
SFDA-DE~\cite{ding2022source} & $\checkmark$ & 59.7& 79.5& 82.4 & 69.7 &  78.6 & 79.2                        &  66.1 & 57.2 &  82.6 &  73.9 &  60.8& 85.2 & 72.9 \\
C-SFDA~\cite{karim2023c} & $\checkmark$ & 60.3& 80.2& \textbf{82.9} & 69.3 &  80.1 & 78.8                        &  67.3 & 58.1 &  83.4 &  73.6 &  61.3& 86.3 & 73.5 \\ \midrule
Ours                     & $\checkmark$ & 59.9  &  79.6                      & 82.7 &\textbf{70.3}   & \textbf{81.8}  &80.4 &  68.5  & 57.8                       & \textbf{83.5} &72.5   &      59.8                   & 86.0                    & \textbf{73.6} \\\bottomrule
\end{tabular}}
\end{table*}

\subsection{Comparison with State-of-the-arts}
\begin{table}[htbp]
\caption{Effectiveness analysis on contrastive-based method and our methods. ``BS'' and ``Mem'' represent the batch size and peak memory on a single GPU. The running time is measured on 1 Tesla A100 GPU with 40 epochs.}
\centering
\label{effect-analysis}
\resizebox{\linewidth}{!}{
\begin{tabular}{ccccc}
\toprule
DomainNet (Rw$\rightarrow$Cl) & Batch Size & Memory            & Time & Accuracy \\ \hline
AdaConstrast~\cite{chen2022contrastive}  & 128        & \textgreater{}32G & -    & 70.2     \\
C-SFDA~\cite{karim2023c}        & 256        & \textgreater{}64G & -    & 70.8     \\
GPL~\cite{litrico2023guiding}           & 256        & \textgreater{}64G & 3h   & 74.2     \\ \hline
Ours          & 128        & 17G               & 2h   & \textbf{76.9}     \\ 
\bottomrule
\end{tabular}}
\end{table}


\begin{table*}[htbp]
\caption{Accuracy (\%) on large-scale \textbf{DomainNet} dataset (ResNet-50). ``SF'' denotes source-free. We highlight the best results.}
\label{exp_domainnet}
\begin{threeparttable}
\resizebox{\linewidth}{!}{
\begin{tabular}{l|c|ccccccc|c}
\toprule
Method                   & SF                        & Rw$\rightarrow$Cl                          & Rw$\rightarrow$Pt                          & Pt$\rightarrow$Cl                          & Cl$\rightarrow$Sk                          & Sk$\rightarrow$Pt                          & Rw$\rightarrow$Sk                          & Pt$\rightarrow$Rw                          & Avg.                        \\ \midrule
ResNet-50~\cite{he2016deep}                & $\times$                                              & 58.8                        & 62.2                        & 57.7                        & 50.3                        & 52.6                        & 47.3                        & 73.2                        & 57.4                        \\ 
MCC~\cite{jin2020minimum}                & $\times$                                              & 44.8                        & 65.7                       &  41.9                       &  34.9                     &     47.3                    &    35.3                    &    72.4                    &   48.9                     \\
CDAN~\cite{long2018conditional}                & $\times$                                              & 65.0                        &64.9                        &  63.7                       & 53.1                      &  63.4                       &   54.5                     &  73.2                      &  62.5                      \\
GVB~\cite{cui2020gradually}                & $\times$                                              & 68.2                        &69.0                        & 63.2                        & 56.6                      &  63.1                       &  62.2                      &   78.3                     &  65.2                      \\
MME~\cite{saito2019semi}                & $\times$                                              &  70.0                       &  67.7                      &  69.0                       & 56.3                      &  64.8                       & 61.0                       &   76.0                     &  66.4                      \\\midrule
TENT~\cite{wang2020tent}                     & $\checkmark$                           & 58.5                        & 65.7                        & 57.9                        & 48.5                        & 52.4                        & 54.0                        & 67.0                        & 57.7                        \\
G-SFDA~\cite{yang2021generalized} & $\checkmark$                            &63.4&67.5&62.5&55.3&60.8&58.3&75.2&63.3\\
NRC~\cite{yang2021exploiting} & $\checkmark$                           &67.5&68.0&67.8&57.6&59.3&58.7&74.3&64.7\\
SHOT~\cite{liang2020we} & $\checkmark$                           &67.7 &68.4 &66.9 &60.1 &66.1 &59.9 &80.8 &67.1\\
 AdaConstrast~\cite{chen2022contrastive}  & $\checkmark$                                &  70.6 &  69.8 &  69.3 & 58.5 & 66.2 & 60.2 &  80.2 &  67.8 \\ 
 AaD~\cite{yang2022attracting} & $\checkmark$                                &  70.2 &  69.8 &  68.6 & 58.0 & 65.9 & 61.5 &  80.5 &  67.8 \\ 
DaC~\cite{zhang2022divide}\tnote{*}                     & $\checkmark$                                             & 70.0                        & 68.8 &  70.9 &  62.4 & 66.8                        & 60.3                        & 78.6                        & 68.3                        \\
                        
C-SFDA\cite{karim2023c} &  $\checkmark$          &  70.8 & 71.1 &  68.5 &  62.1 &  67.4& 62.7 & 80.4 &  69.0 \\
GPL~\cite{litrico2023guiding}                      & $\checkmark$                             & 74.2 &  70.4 &  68.8 & 64.0 & 67.5 & \textbf{65.7} &  76.5 & 69.6 \\  \midrule
Ours                     & $\checkmark$                          & \textbf{76.9} & \textbf{71.8} & \textbf{75.4} &\textbf{65.5}  & \textbf{69.9} & 64.6 & \textbf{83.2}  &  \textbf{72.5}\\\bottomrule
\end{tabular}}
\begin{tablenotes}
\footnotesize
    \item[*] This work uses ResNet-34 as backbone.
\end{tablenotes}
\end{threeparttable}
\end{table*}

\begin{table*}[htbp]
\caption{Accuracy (\%) on large-scale \textbf{VisDA-C} dataset (ResNet-101). ``SF'' denotes source-free. We highlight the best results.}
\resizebox{\linewidth}{!}{
\label{exp_visda-c}
\begin{tabular}{l|c|cccccccccccc|c}
\toprule
Method                   &SF              & plane                       & bcyle                       & bus                         & car                         & horse                       & knife                       & mcyle                       & person                      & plant                       & sktbrd                      & train                       & truck                       & Avg.                        \\ \midrule
ResNet-101~\cite{he2016deep}               & $\times$         & 55.1                        & 53.3                        & 61.9                        & 59.1                        & 80.6                        & 17.9                        & 79.7                        & 31.2                        & 81.0                        & 26.5                        & 73.5                        & 8.5                         & 52.4                        \\
MCC~\cite{jin2020minimum}               & $\times$         & 88.7                    &  80.3                       & 80.5                        &   71.5                    & 90.1                       &93.2                       & 85.0                      &  71.6                       & 89.4                        &  73.8                      &  85.0                       &   36.9                       &   78.8                      \\
STAR~\cite{lu2020stochastic}               & $\times$         & 95.0                    &  84.0                       &  84.6                       & 73.0                      &91.6                        &  91.8                     & 85.9                      & 78.4                        &94.4                         & 84.7                       &  87.0                       &  42.2                        &   82.7                      \\
RWOT~\cite{xu2020reliable}               & $\times$         & 95.1                    & 87.4                        & 85.2                        &  58.6                     &   96.2                     &  95.7                     & 90.6                      &  80.0                       & 94.8                        &   90.8                     &   88.4                      & 47.9                         &  84.3                       \\
CAN~\cite{kang2019contrastive}               & $\times$         & 97.0                    & 87.2                        & 82.5                        &  74.3                     & \textbf{97.8}                      & 96.2                      & 90.8                      &  80.7                       &   96.6                      &  \textbf{96.3}                   &  87.5                       &  59.9                        &   87.2                    \\\midrule
SHOT~\cite{liang2020we}                     & $\checkmark$              & 94.3                        & 88.5                        & 80.1                        & 57.3                        & 93.1                        & 94.9                        & 80.7                        & 80.3                        & 90.5                        & 89.1                        & 86.3                        & 58.2                        & 82.9                        \\
DIPE~\cite{wang2022exploring}                     & $\checkmark$               & 95.2                        & 87.6                        & 78.8                        & 55.9                        & 93.9                        & 95.0                        & 84.1                        & 81.7                        & 92.1                        & 88.9                        & 85.4                        & 58.0                        & 83.1                        \\
HCL~\cite{huang2021model}                      & $\checkmark$              & 93.3                        & 85.4                        & 80.7                        & 68.5                        & 91.0                        & 88.1                        & 86.0                        & 78.6                        & 86.6                        & 88.8                        & 80.0                        & \textbf{74.7} & 83.5                        \\ 
$A^2$Net~\cite{xia2021adaptive}& $\checkmark$              & 94.0                        &  87.8                       &85.6                        &  66.8                     & 93.7                       & 95.1                       &  85.8                      & 81.2                        & 91.6                        & 88.2                        &  86.5                      &56.0  &84.3                         \\ 
G-SFDA~\cite{yang2021generalized}                   &$\checkmark$             & 96.1                        & 88.3                        & 85.5                        & 74.1                        & 97.1                        & 95.4                        & 89.5                        & 79.4                        & 95.4                        & 92.9                        & 89.1                        & 42.6                        & 85.4                        \\
NRC~\cite{yang2021exploiting}                      & $\checkmark$               & 96.8                        &  \textbf{91.3} & 82.4                        & 62.4                        & 96.2                        & 95.9                        & 86.1                        & 80.6                        & 94.8                        & 94.1                        & 90.4                        & 59.7                        & 85.9                        \\
SFDA-DE~\cite{ding2022source}                & $\checkmark$            & 95.3                        & 91.2                        & 77.5                        & 72.1                        & 95.7                        & \textbf{97.8}                        & 85.5                        &  86.1 & 95.5                        & 93.0                        & 86.3                        & 61.6                        & 86.5                        \\
AdaContrast~\cite{chen2022contrastive} & $\checkmark$ & 97.0                        & 84.7                        & 84.0                        & 77.3                        & 96.7                        & 93.8                        & 91.9                        & 84.8                         & 94.3                        & 93.1                        &  \textbf{94.1} & 49.7                        & 86.8                        \\
CoWA~\cite{lee2022confidence}                  & $\checkmark$             & 96.2                        & 89.7                       & 83.9                        & 73.8                        & 96.4                        & 97.4                        & 89.3                        & \textbf{86.8} & 94.6                        & 92.1                       & 88.7                        & 53.8                        & 86.9                        \\
DaC~\cite{zhang2022divide}    & $\checkmark$              & 96.6                        & 86.8                        & \textbf{86.4} &  78.4 & 96.4                        & 96.2 &  \textbf{93.6} & 83.8                        & \textbf{96.8} &  95.1 & 89.6                        & 50.0                        & 87.3                        \\
BDT~\cite{kundu2022balancing}& $\checkmark$  &  - & -                        & -                        & -                        & - & -                       & -                        & -                        & -                        & -                        & - &  - &  87.8 \\
C-SFDA~\cite{karim2023c} & $\checkmark$  &  97.6 & 88.8                        & 86.1                        & 72.2                        & 97.2 & 94.4                        & 92.1                        & 84.7                        & 93.0                        & 90.7                        & 93.1 &  63.5 &  87.8 \\ \midrule
Ours                     & $\checkmark$               & \textbf{98.0} & 88.0                       & \textbf{86.4} &  \textbf{82.3} &  \textbf{97.8} &  96.2                         & 92.1                        & 85.0 & 95.5 & 91.7        &93.8               & 56.2                        & \textbf{88.6} \\
\bottomrule
\end{tabular}}
\end{table*}


\subsubsection{Quantitative Results}
We compare our proposed method against popular source-present and source-free methods on three benchmark datasets: Office-Home, DomainNet, and VisDA-C. We report the Top-1 accuracy, and the results are presented in Table~\ref{exp_office-home} to Table~\ref{exp_visda-c}. In the Office-Home dataset, as shown in Table~\ref{exp_office-home}, our proposed method achieves the best performance in terms of Top-1 average accuracy, which is comparable to the most recent source-free method C-SFDA. Additionally, our method in 3 sub-transfer tasks achieves the highest accuracy (see bold in Table~\ref{exp_office-home}) vs. only one sub-transfer task in C-SFDA. For the DomainNet dataset, as demonstrated in Table~\ref{exp_domainnet}, our proposed method exhibits significant improvements over all baselines. With an average Top-1 accuracy of $72.5\%$, our method outperforms the best source-free baseline by nearly 3\% and surpasses the best source-present baseline by 6.1\%. Moreover, our method achieves the best performance in almost all domain shifts. On the VisDA-C dataset, presented in Table~\ref{exp_visda-c}, our proposed method outperforms the state-of-the-art method C-SFDA~\cite{karim2023c} by 0.8\%. Furthermore, our method achieves the best performance in specific classes such as ``plane'', ``bus'', ``car'', and ``horse''. These results clearly demonstrate the superiority of our proposed method across the evaluated datasets, showcasing its effectiveness in source-free domain adaptation scenarios.

\subsubsection{Effectiveness Analysis} 
We conducted an analysis and comparison of the memory usage and running time of our method with recent works, including AdaContrast~\cite{chen2022contrastive}, C-SFDA~\cite{karim2023c}, and GPL~\cite{litrico2023guiding}. Interestingly, our method requires normal memory usage, whereas the other methods consume more than 32GB of memory. Despite using standard memory, our approach achieves higher accuracy in comparison. Additionally, the running time of our method is considerably less than that of GPL.

\subsection{Ablation Studies}

\begin{table}[htbp]
    \caption{\textbf{DomainNet} (Pt$\rightarrow$Cl) Top-1 accuracy (\%) of the proposed method with the different number of the prediction hypotheses $\tilde{k}$. We find $\tilde{k}=4$ yields the optimal results.}
\label{ablation_k}
\centering
    \begin{tabular}{ccccccc}
\toprule
 $\tilde{k}$ & 2&3&4&5&6  \\ \midrule
        Accuracy       & 73.7   & 74.2 & \textbf{75.4}  & 75.3 &74.8    \\\bottomrule     
\end{tabular} 
\end{table}

\subsubsection{Component-wise Analysis} 
In this section, we conduct ablation studies to analyze the contribution of each component in our method on three benchmark datasets: Office-Home, DomainNet, and VisDA-C. The results are summarized in Table~\ref{ablation_component}, in which the HCPR (Hypothesis Consolidation from Prediction Rationale) component makes the most contributions to the promotion of accuracy. Specifically,  compared to only using FixMatch, combining both FixMatch and HCPR significantly improves accuracy by 4.4\%, 10.0\%, and 22.9\% on the respective datasets. Additionally, in the case of combining both PA (Pre-Adaptation) and HCPR, we execute PA again following HCPR to integrate the consolidation outcomes from HCPR. This showcases a substantial enhancement in accuracy, with improvements of 0.6\%, 2.2\%, and 1.3\% on the respective datasets compared to solely employing PA. Last but not least, Removing HCPR from the method leads to a performance drop of 1.4\%, 5\%, and 2\% points on Office-Home, DomainNet, and VisDA-C, respectively.

\begin{figure}[tbp]
	\centering
	\includegraphics[width=3.5in]{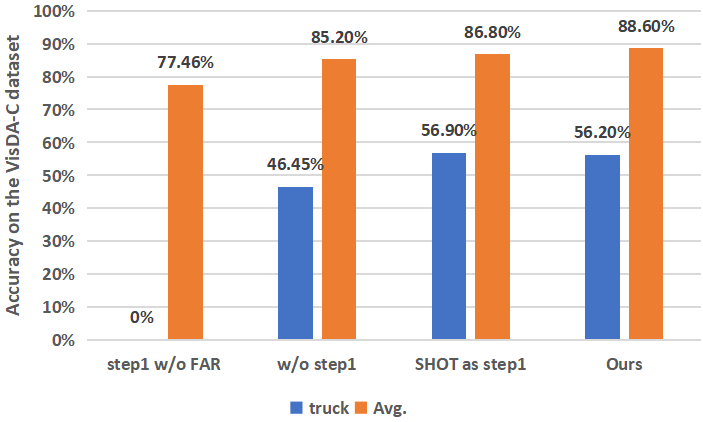}
	\caption{Comparison of step 1 w/o FAR, w/o step 1, SHOT as step 1, and Ours on the trunk accuracy and average top-1 accuracy on the \textbf{VisDA-C} dataset. }
	\label{visual_PLG}
\end{figure}

\subsubsection{Impact of Model Pre-adaptation} 
To assess the impact of model pre-adaptation on per-class accuracy (e.g., "truck") and average accuracy in our method, we perform experiments using four different settings on the VisDA-C dataset: model pre-adaptation removing the second term $\mathcal{L}_{FAR}$ in Eq. 1 referring to ``step 1 w/o FAR''; the proposed method without model pre-adaptation referring to ``w/o step 1''; Using SHOT's loss as model pre-adaptation to replace Eq. 1, referring to ``SHOT as step 1''; and the proposed method with model pre-adaptation using Eq. 1 referring to ``Ours''. The experimental results are shown in Figure~\ref{visual_PLG}. As we can see, we have the following observations: First, compared to ``SHOT as step 1'', the proposed method encouraging smooth prediction has a better average accuracy (86.80\% vs. 88.60\%), which demonstrates the superiority of making a smooth prediction on the data manifold compared to one-hot prediction in~\cite{liang2020we}. Second, when removing step 1, referring to ``w/o step 1,'' the average accuracy dropped by 3.4\%. This indicates that Eq. 1 is helpful for model pre-adaptation and improves the ability of the model to distinguish image classes in the target domain. Third, when removing $\mathcal{L}_{FAR}$ in step 1 referring to ``step 1 w/o FAR'', the average performance drop dramatically from 88.60\% to 77.46\% and the classification accuracy in the class ``trunk'' drop to 0\%. This demonstrates that the $\mathcal{L}_{FAR}$ plays a vital role in keeping class balance and avoiding some missed classes. 

\subsubsection{Impact of $\tilde{k}$ \textemdash the Number of Prediction Hypotheses Per Instance} 
In our method, we choose labels from the top $\tilde{k}$ highest posterior probabilities as the prediction hypothesis. In this section, we investigate the impact of the value of $\tilde{k}$. Table~\ref{ablation_k} shows the accuracy achieved with different $\tilde{k}$. From the result, we can see that using 2 hypotheses has already led to good performance and choosing 3-6 hypotheses leads to optimal performance.


\begin{figure}[tbp]
	\centering
	\includegraphics[width=3.4in]{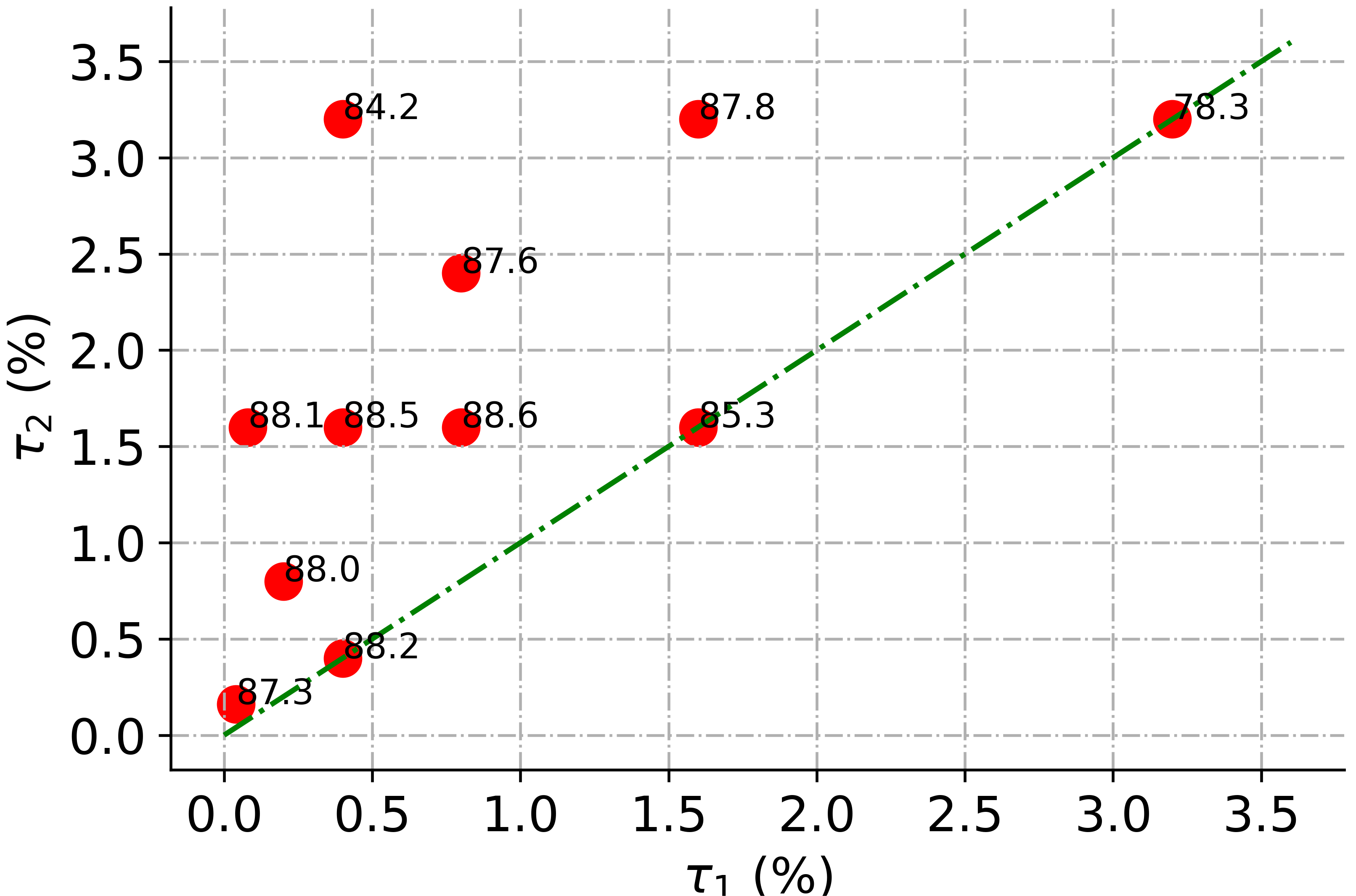}
	\caption{\textbf{VisDA-C} average accuracy (\%) of the proposed method using different $\tau_1$ and $\tau_2$.}
	\label{u1_u2}
 \end{figure}
 
\begin{table*}[htbp]
\caption{Accuracy (\%) of our method combined with existing SHOT and AaD methods on the \textbf{Office-Home}, \textbf{VisDA-C} and \textbf{DomainNet} datasets.}
\centering
\label{intergation_SHOT}
\resizebox{\linewidth}{!}{
\begin{tabular}{l|cccccccccccc|c}
\toprule
Method                    & Ar$\rightarrow$Cl                        & Ar$\rightarrow$Pr                        & Ar$\rightarrow$Rw                        & Cl$\rightarrow$Ar                        & Cl$\rightarrow$Pr                        & Cl$\rightarrow$Rw                        & Pr$\rightarrow$Ar                        & Pr$\rightarrow$Cl                        & Pr$\rightarrow$Rw                        & Rw$\rightarrow$Ar                        & Rw$\rightarrow$Cl                        & Rw$\rightarrow$Pr                        & Avg.                        \\ \midrule
SHOT~\cite{liang2020we}                       & 56.9                        & 78.1                        & 81.0                        & 67.9                        & 78.4                        & 78.1                        & 67.0                        & 54.6                        & 81.8                        & 73.4                        & 58.1                        & 84.5                        & 71.6                        \\
SHOT+Ours                     &58.7   & \textbf{79.5}                     & 82.1 &69.6   &80.7   &\textbf{80.0} & \textbf{69.1}   &56.9                        &82.3  &\textbf{74.5}  &59.2                         & 85.3                   &73.2  \\
AaD~\cite{yang2022attracting} & 59.3                        & 79.3                        & 82.1                        & 68.9                        & 79.8                       & 79.5                        & 67.2                        & 57.4                        & 83.1                        & 72.1                        & 58.5                        & 85.4                        & 72.7                        \\
AaD+Ours & \textbf{59.8}                        & 79.4                        & \textbf{82.7 }                       & \textbf{70.0}                        & \textbf{81.6 }                      & \textbf{80.0}                        & 68.5                        & \textbf{57.6}                        & \textbf{83.2}                        & 72.7                        & \textbf{59.4}                        & \textbf{86.1}                        &   \textbf{73.4}                      \\
\bottomrule
\end{tabular}}

\resizebox{\linewidth}{!}{
\begin{tabular}{l|cccccccccccc|c}
\toprule
Method                            & plane                       & bcyle                       & bus                         & car                         & horse                       & knife                       & mcyle                       & person                      & plant                       & sktbrd                      & train                       & truck                       & Avg.                        \\ \midrule
SHOT~\cite{liang2020we}                             & 94.3                        & 88.5                        & 80.1                        & 57.3                        & 93.1                        & 94.9                        & 80.7                        & 80.3                        & 90.5                        & 89.1                        & 86.3                        & 58.2                        & 82.9                        \\
SHOT+Ours                                &97.5  &84.6                        &83.0  &74.2   &96.5   & 93.7                          & 92.8                       & \textbf{86.7} &93.5  & 92.6        &  89.7            & 56.9                        &86.8  \\
AaD~\cite{yang2022attracting} &97.4  &\textbf{90.5}                        &80.8  &76.2   &97.3   & \textbf{96.1}         & 89.8                       & 82.9 &\textbf{95.5}  & \textbf{93.0}        &  92.0            & \textbf{64.0}                        &88.0  \\
AaD+Ours &\textbf{97.8}  &87.6                        &\textbf{86.7}  &\textbf{83.4}  &\textbf{97.7}   & 95.4        &\textbf{94.2} & 83.8                       & 94.6 &91.2       &  \textbf{92.8}           & 55.6                        &\textbf{88.4}  \\
\bottomrule
\end{tabular}}

\setlength{\tabcolsep}{4mm}{
\begin{tabular}{l|ccccccc|c}
\toprule
Method                                           & Rw$\rightarrow$Cl                          & Rw$\rightarrow$Pt                          & Pt$\rightarrow$Cl                          & Cl$\rightarrow$Sk                          & Sk$\rightarrow$Pt                          & Rw$\rightarrow$Sk                          & Pt$\rightarrow$Rw                          & Avg.                        \\ \midrule
SHOT~\cite{liang2020we}                           &67.7 &68.4 &66.9 &60.1 &66.1 &59.9 &80.8 &67.1\\
SHOT+Ours                                             &70.5  & 70.6 &72.5  &63.6  &68.0  & 61.1 & \textbf{82.8}  &69.9 \\
AaD~\cite{yang2022attracting}&70.6  & 69.8 &69.3  &58.5  &66.2  & 60.2 & 80.2  &67.8 \\
AaD+Ours &\textbf{75.4}  & \textbf{71.3} &\textbf{75.2} &\textbf{64.2}  &\textbf{68.4}  & \textbf{63.3} & \textbf{82.8}  &\textbf{71.5} \\
\bottomrule
\end{tabular}}
\end{table*}

\begin{figure}[htbp]
	\centering
	\includegraphics[width=3.4in]
 {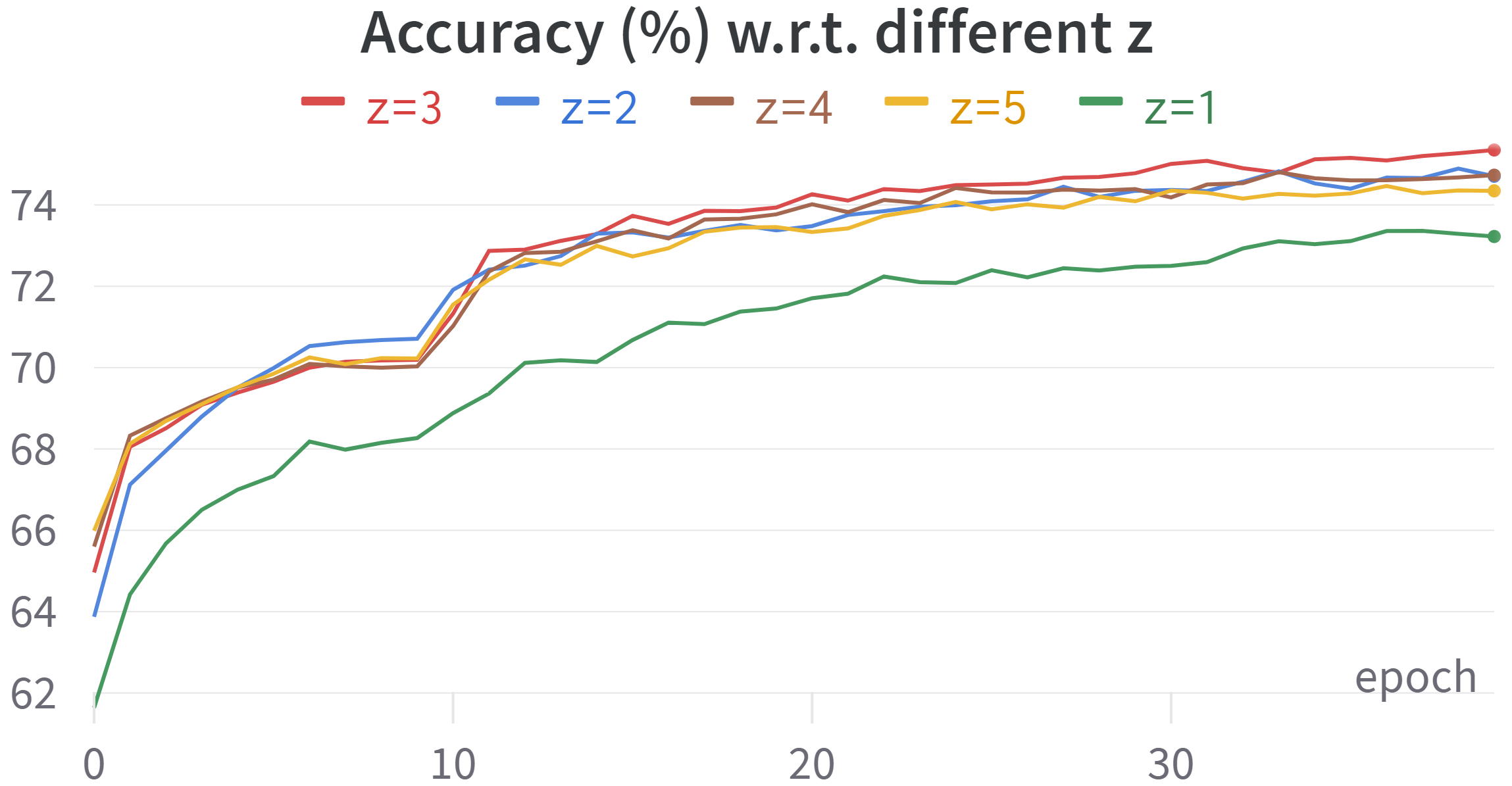}
	\caption{Accuracy of different $z$ in our method on the \textbf{DomainNet} (Pt$\rightarrow$Cl) dataset. When step 1 (0-9 epochs) achieves and maintains the best results, HCPR plays a pivotal role in enhancing the performance of the model.}
	\label{appendix_z}
\end{figure}
\begin{table}[htbp]
    \caption{\textbf{DomainNet} (Pt$\rightarrow$Cl) accuracy (\%) of the proposed method with different number of the $z$.}
\label{ablation_z}
\centering
    \begin{tabular}{cccccc}
\toprule
 $z$ & 1&2&3&4&5  \\ \midrule
        Accuracy       & 73.2   & 74.7 & \textbf{75.4}  & 74.7  & 74.3      \\\bottomrule     
\end{tabular} 
\end{table}

\subsubsection{Impact of the $z$ \textemdash the Number of Nearest and Furthest Neighbor} 
In the initial step of our model pre-adaptation, we select the $z$-nearest and $z$-furthest neighbors for each target sample. In this analysis, we examine the influence of the $z$ value. Figure~\ref{appendix_z} and Table~\ref{ablation_z}  showcase the performance throughout the training process and top-1 accuracy of classification on the DomainNet (Pt$\rightarrow$Cl) dataset for different values of $z$. The results indicate that even with just one nearest and furthest neighbor, we achieve favorable classification accuracy, and selecting 2-5 nearest and furthest neighbors yields optimal performance. Moreover, as observed in Figure~\ref{appendix_z}, it is worth noting that when step 1 (0-9 epochs) achieves and maintains the best results, HCPR plays a pivotal role in enhancing the performance of the model.

\subsubsection{Impact of the Two Ranking Thresholds $\tau_1$ and $\tau_2$} 
To assess the influence of ranking thresholds in our method, we examined the percentage values $\tau_1$ and $\tau_2$ relative to the total number of samples. Specifically, we analyzed their impact on the Top-1 average accuracy on the VisDA-C dataset, as illustrated in Figure~\ref{u1_u2}. Our analysis, depicted in Figure~\ref{u1_u2}, revealed that the proposed method exhibits robustness to the specific values of $\tau_1$ and $\tau_2$.

\begin{table}[htbp]
    \caption{Average accuracy (\%) of our HCPR vs. near centroid collection on the \textbf{Office-Home} and \textbf{DomainNet} datasets.}
\label{replaceable}
\centering
    \begin{tabular}{ccc}
\toprule
 Method & Office-Home & DomainNet \\ \midrule
        near-centroid selection        & 72.6      & 69.6    \\
        Ours        & \textbf{73.6}      & \textbf{72.5}    \\\bottomrule
\end{tabular} 
\end{table}

 \begin{figure}[htbp]
	\centering
	\includegraphics[width=3.0in]{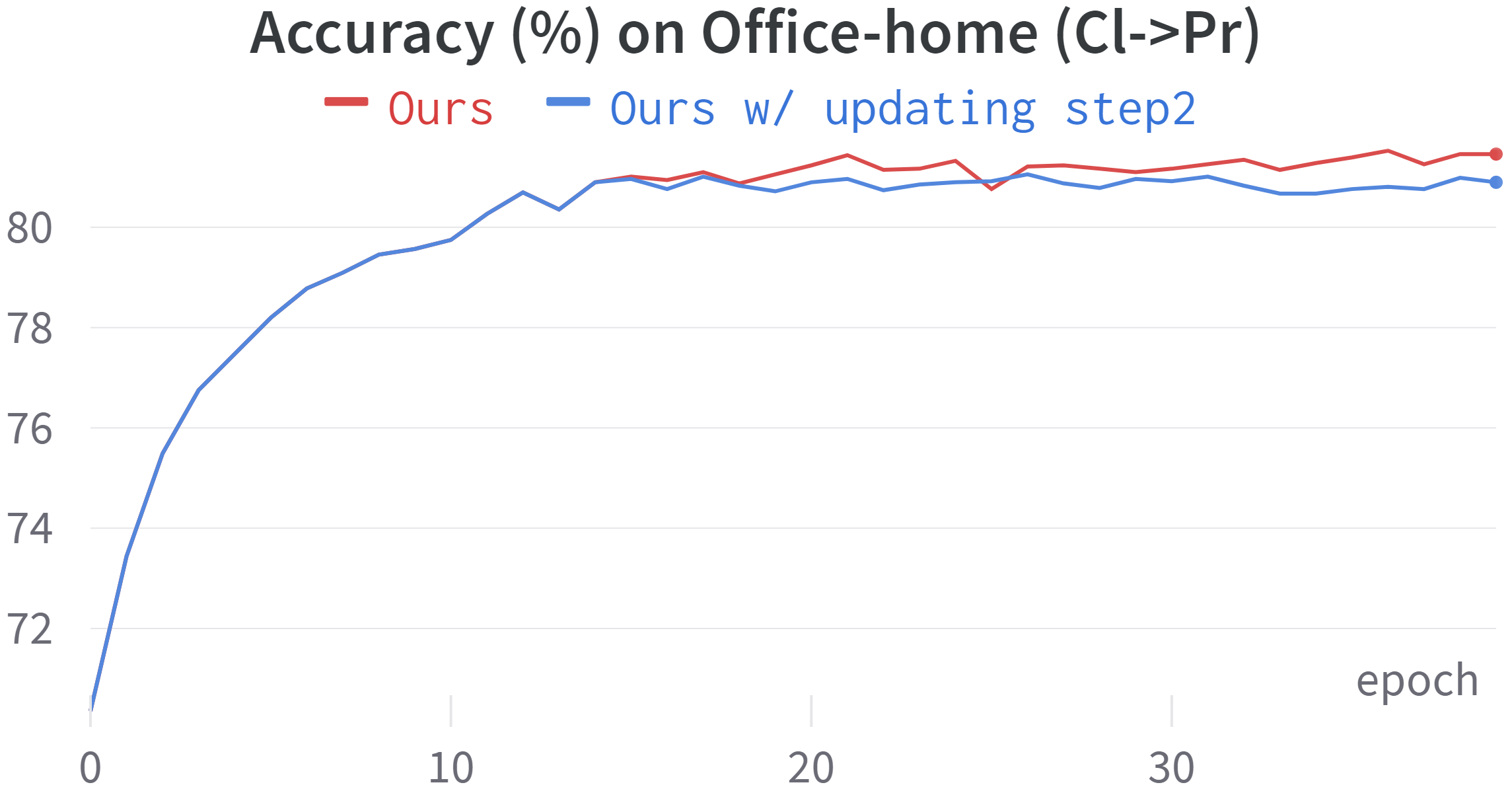}
	\caption{\textbf{Office-Home} accuracy of Ours and Ours w/ updating step 2 across varying epochs.}
	\label{visual_update_step2}
 \end{figure}
\subsubsection{The Benefit of Using Rationale Representations} 
To further understand the benefit of using the rationale representation from multiple hypotheses, we explore an alternative method that replaces the proposed second step by using feature centroids rather than rationale centroids. Since the feature is invariant to the prediction hypothesis, only the top predicted class will be considered. More specially, we first generate pseudo-label for each instance and calculate the feature centroid similar to our approach. Then we rank instances based on the Euclidean distances between their features and the corresponding class centroid. 
The top $\tau_1$ features closest to the class centroid are assigned reliable pseudo labels, while the remaining samples are left for step 3. We refer to this method as ``near-centroid selection". Table~\ref{replaceable} presents the comparison results on the Office-Home and DomainNet datasets. As seen, while such an approach still leads to improvement over using step 1 and step 3 alone (by cross-referencing Table~\ref{ablation_component}), it is still inferior to the use of HCPR. This clearly demonstrates the benefits of the latter.


\subsubsection{Investigation of Recursively Applying HCPR} 
One may wonder if recursively applying HCPR will lead to additional improvement. To this end, we create a variant of our method by alternatively applying step 2 and step 3, hoping that they may mutually enhance each other. 
We conducted experiments on the Office-Home (Cl$\rightarrow$Pr) dataset. The results are depicted in Figure~\ref{visual_update_step2}, where the red curve represents our method using the second step only once, i.e., the hypothesis consolidation occurs between model pre-adaptation (0-9 epochs) and semi-supervised learning (10-40 epochs). The blue curve represents our method with the second step updated at the 15th, 20th, and 25th epochs. From the results, we observed that recursively applying HCPR does not lead to an improvement as one may expect. 

\begin{figure}[tbp]
	\centering
	\includegraphics[width=3.4in]
 {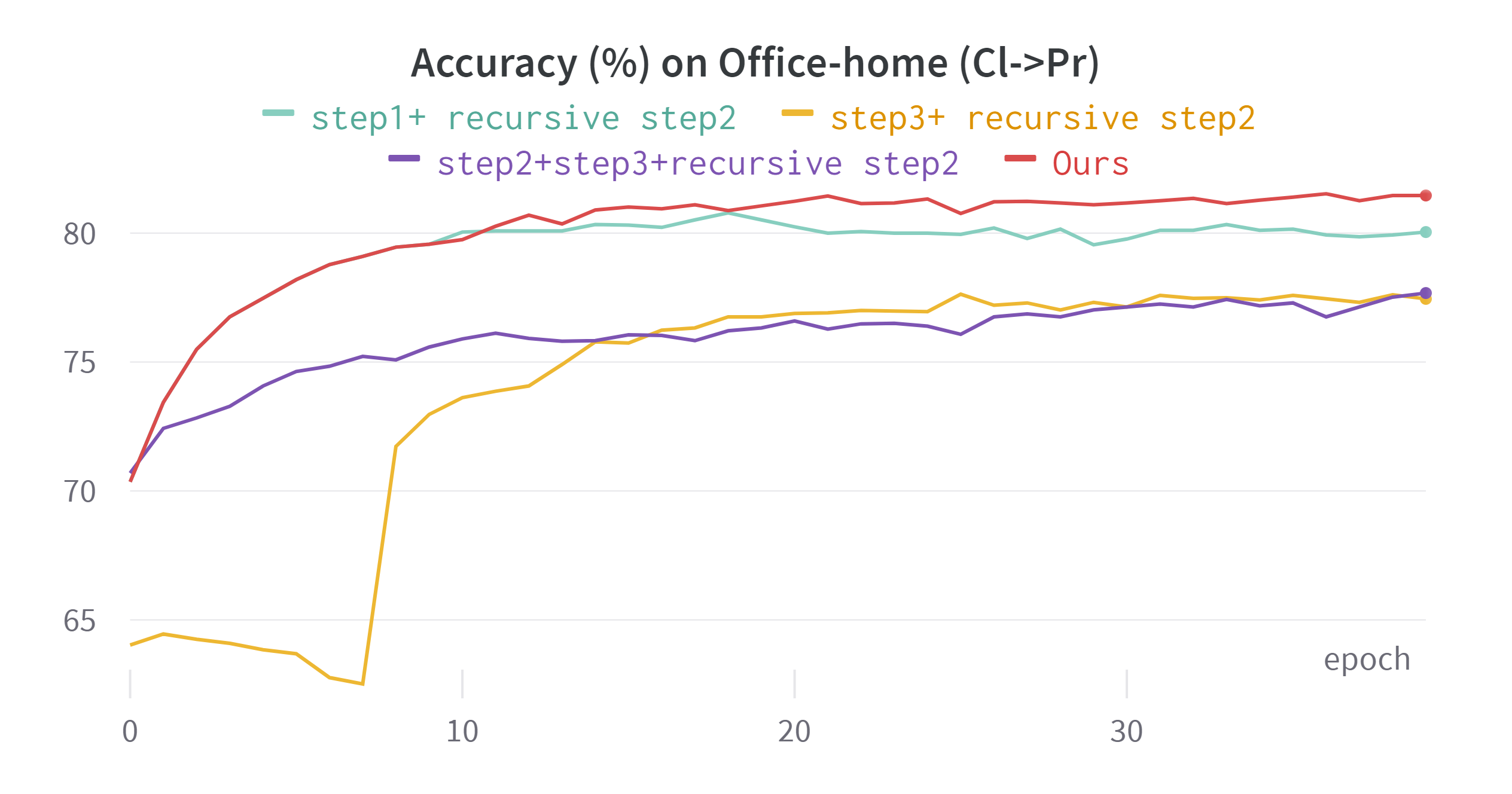}
	\caption{Accuracy of different components in our method with Recursive HCPR on the Office-Home (Cl$\rightarrow$Pr) dataset.}
	\label{appendix_recursive_HCPR}
\end{figure}
We also conduct experiments with HCPR applied recursively to only model pre-adaptation or FixMatch. Specifically, we conduct experiments using the Office-Home (Cl$\rightarrow$Pr) dataset and configure the following scenarios:
\begin{itemize}
\item Combining Step 1 and Step 2, with Step 2 calculated at the 9th, 15th, and 20th epochs (indicated by the green curve in Figure~\ref{appendix_recursive_HCPR}).
\item Combining Step 3 and Step 2, with Step 2 calculated at the 7th, 15th, and 20th epochs (indicated by the yellow curve in Figure~\ref{appendix_recursive_HCPR}).
\item Combining Step 2 and Step 3, with Step 2 calculated at the 0th, 15th, and 20th epochs (indicated by the purple curve in Figure~\ref{appendix_recursive_HCPR}).
\item Our method, is represented by the red curve.
\end{itemize}

Our observations indicate that utilizing Step 2 only once is sufficient, and the recursive HCPR application does not yield improvements. However, we do note that HCPR plays a crucial role in enhancing FixMatch, particularly in improving the quality of pseudo-labels.

\subsection{Pseudo-label Quantity and Quality}
\label{app_qq}
\begin{table}[htbp]
\caption{Comparison of pseudo-label quantity and quality on \textbf{DomainNet} (Rw$\rightarrow$Cl). Quantity (\%) refers to the proportion of selected samples to total samples. Quality (\%) refers to the precision (\%) of the chosen sample. ``con'' represents confidence.}
\label{appendix_qq}
\centering
\resizebox{\linewidth}{!}{
\begin{tabular}{ccc}
\toprule
DomainNet (Rw-\textgreater{}Cl) & Quantity (\%)                                            & Quality (\%)              \\ \hline
source model only (con\textgreater{}0.95) & 3.95                                                 & 95.80                            \\
SHOT (con\textgreater{}0.95)              & 61.83                                                & 80.38                            \\
PA only (con\textgreater{}0.95)           & 79.13                                                & 80.76                            \\
HCPR only                               & 21.35                                                & 84.02                            \\
PA+HCPR                            & 24.65                                                & 90.76                            \\ \bottomrule
\end{tabular}}
\end{table}

\begin{figure*}[htbp]
	\centering
	\subfigure[Pseudo-label quantity and quality w.r.t. different epochs.] 
	{
		\begin{minipage}{7.0cm}
			\centering          
			\label{qq_course_train}
			\includegraphics[scale=0.055]{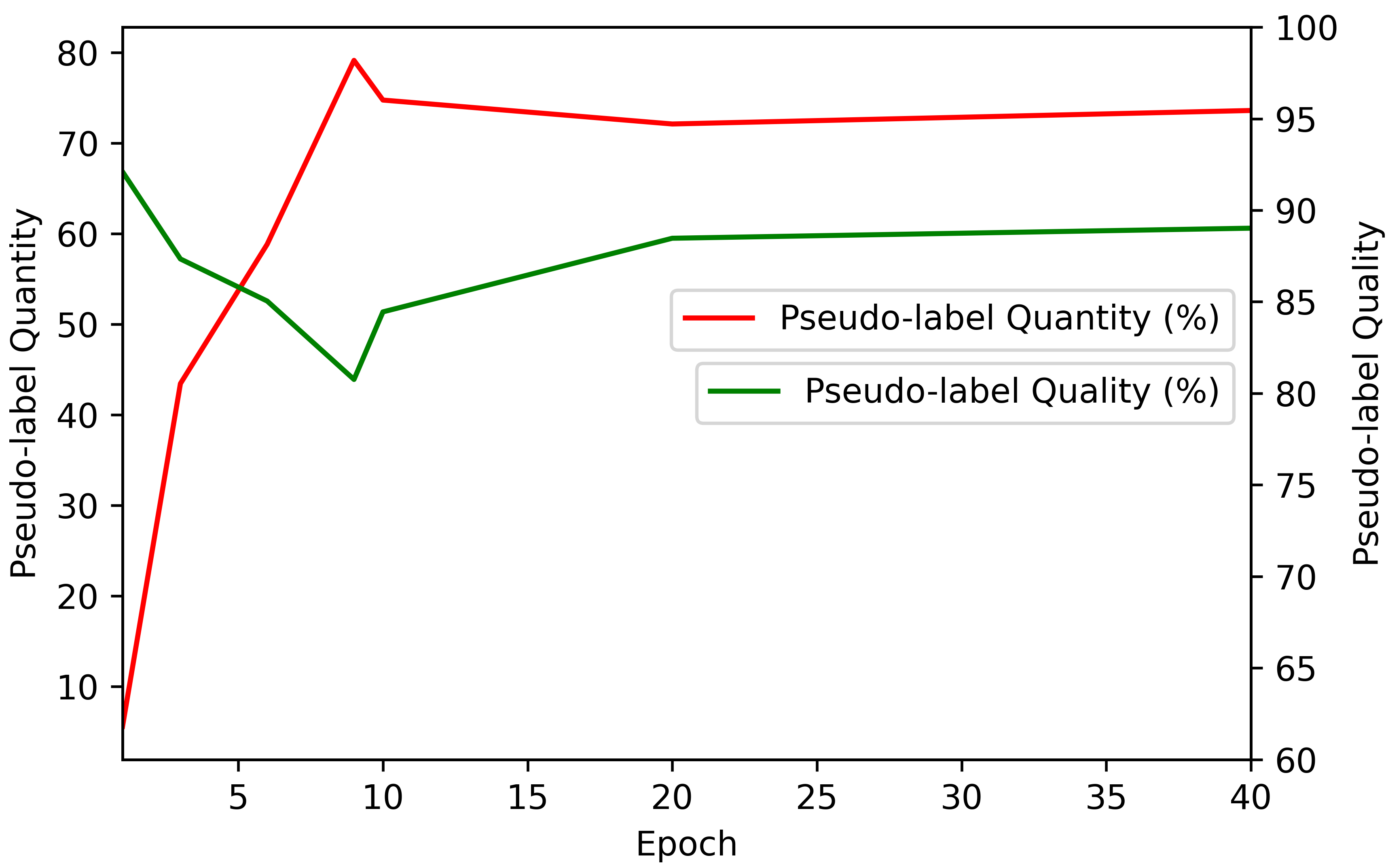}   
		\end{minipage}
	}
	\subfigure[Classification accuracy w.r.t. different epochs.] 
	{
		\begin{minipage}{7.0cm}
			\centering          
			\label{acc_course_train}
			\includegraphics[scale=0.055]{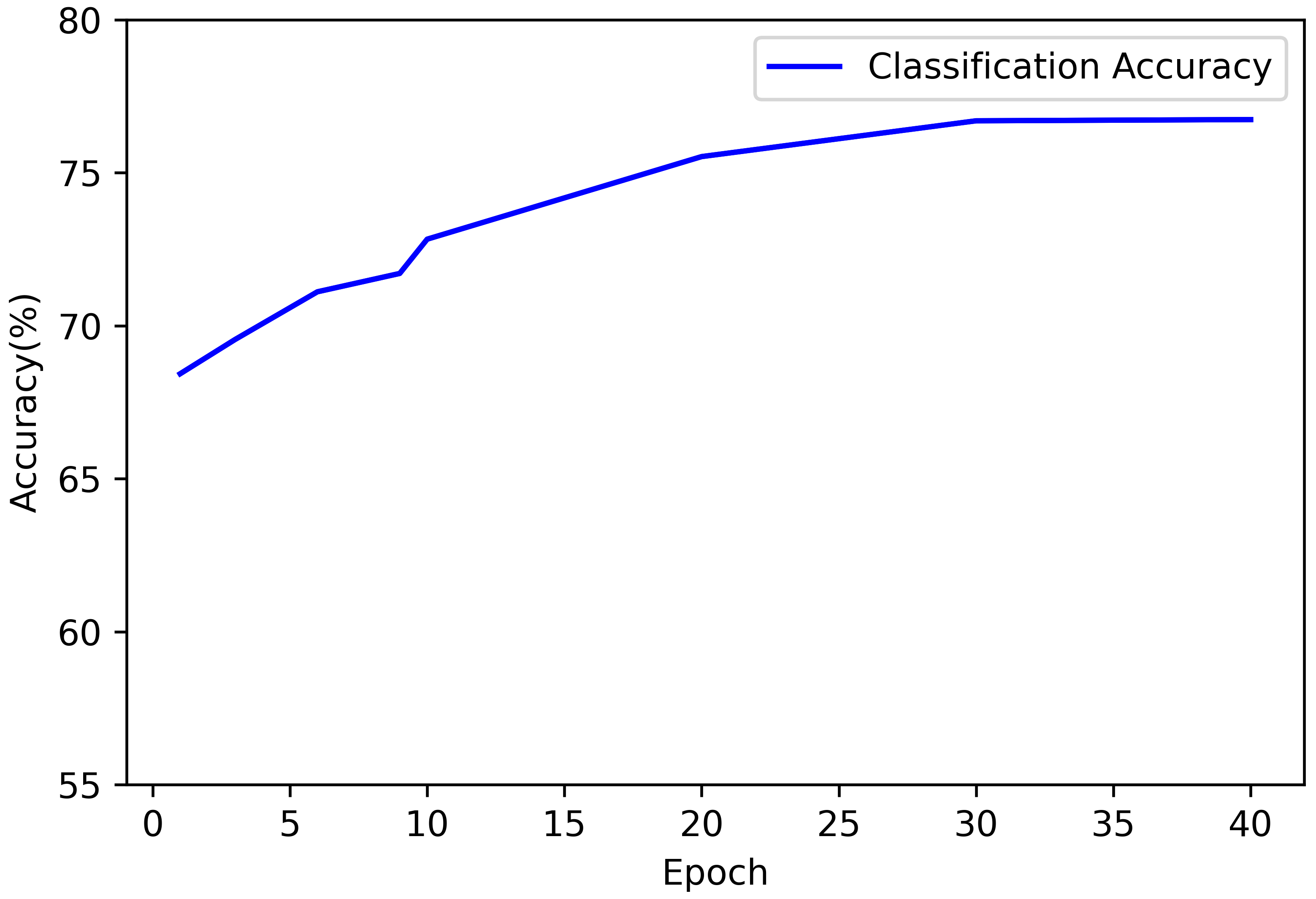}   
		\end{minipage}
	}
	\caption{Pseudo-label quantity, quality and classification accuracy of our method over training on DomainNet (Rw$\rightarrow$Cl).}   
	\label{appendix_course}  
\end{figure*}

\begin{figure*}[htbp]
	\centering
	\subfigure[Source model only] 
	{
		\begin{minipage}{4.5cm}
			\centering          
			\label{a}
			\includegraphics[scale=0.035]{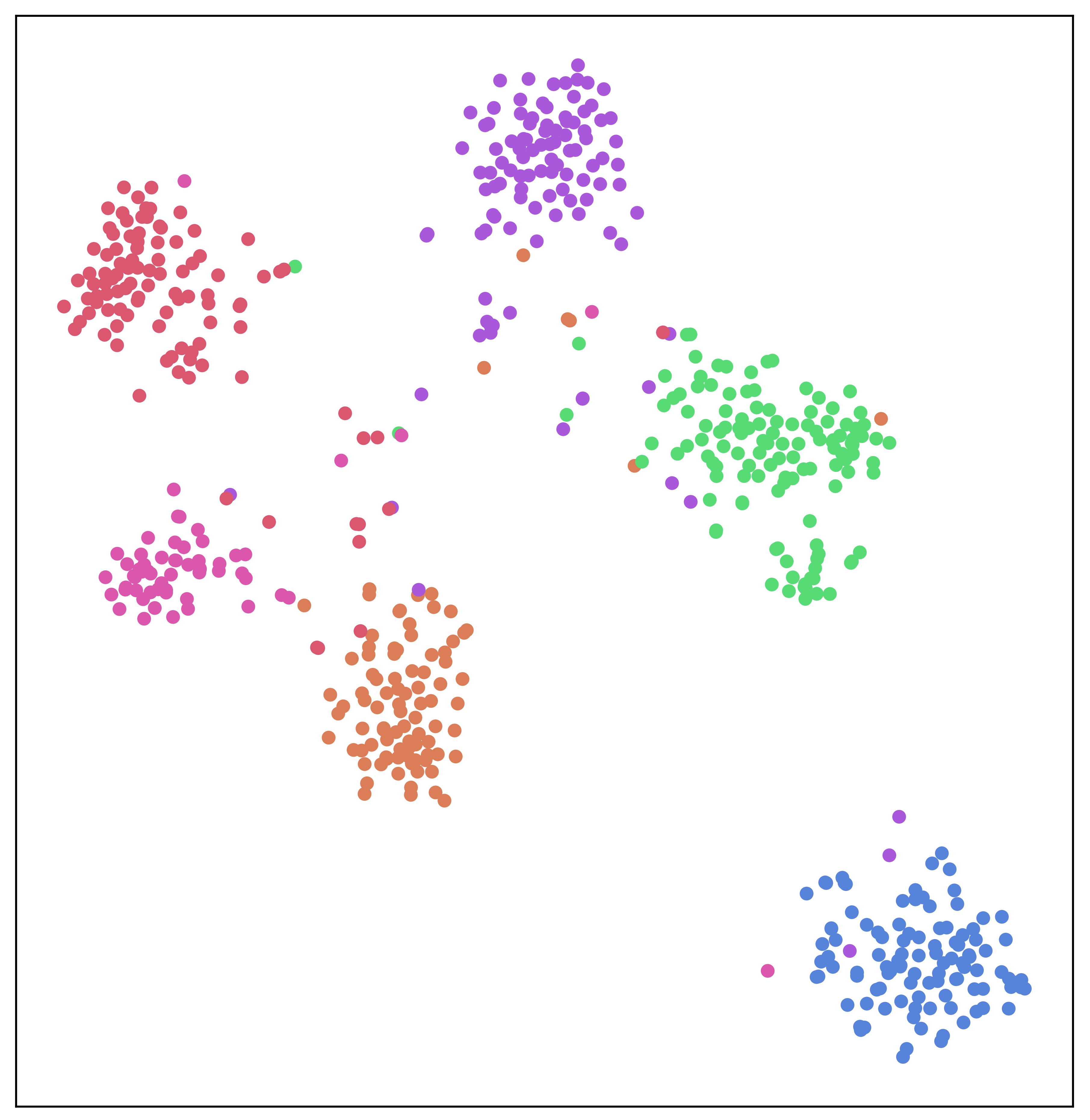}   
		\end{minipage}
	}
	\subfigure[AaD] 
	{
		\begin{minipage}{4.5cm}
			\centering          
			\label{b}
			\includegraphics[scale=0.035]{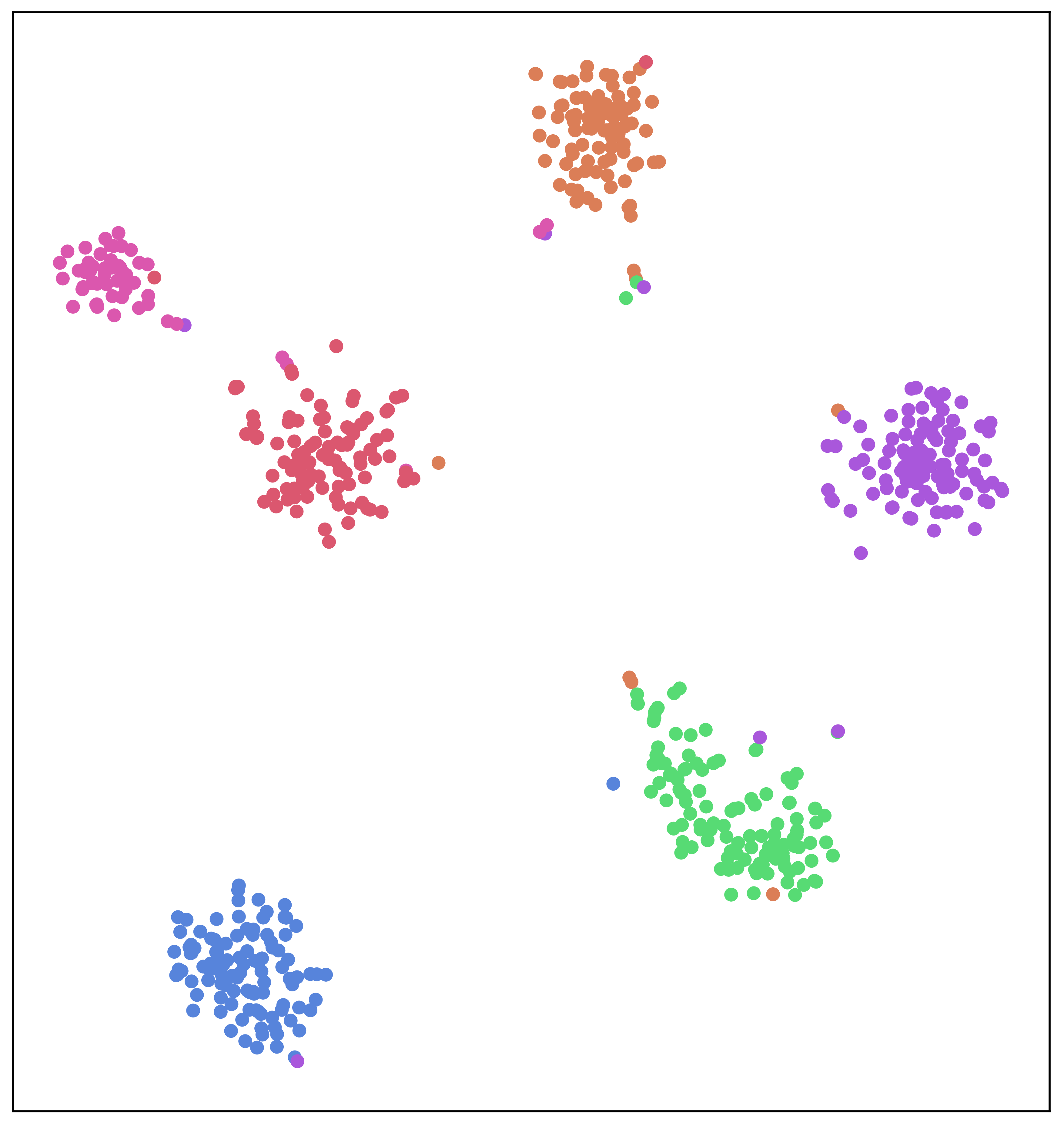}   
		\end{minipage}
	}
	\subfigure[Ours] 
	{
		\begin{minipage}{4.5cm}
			\centering          
			\label{c}
			\includegraphics[scale=0.035]{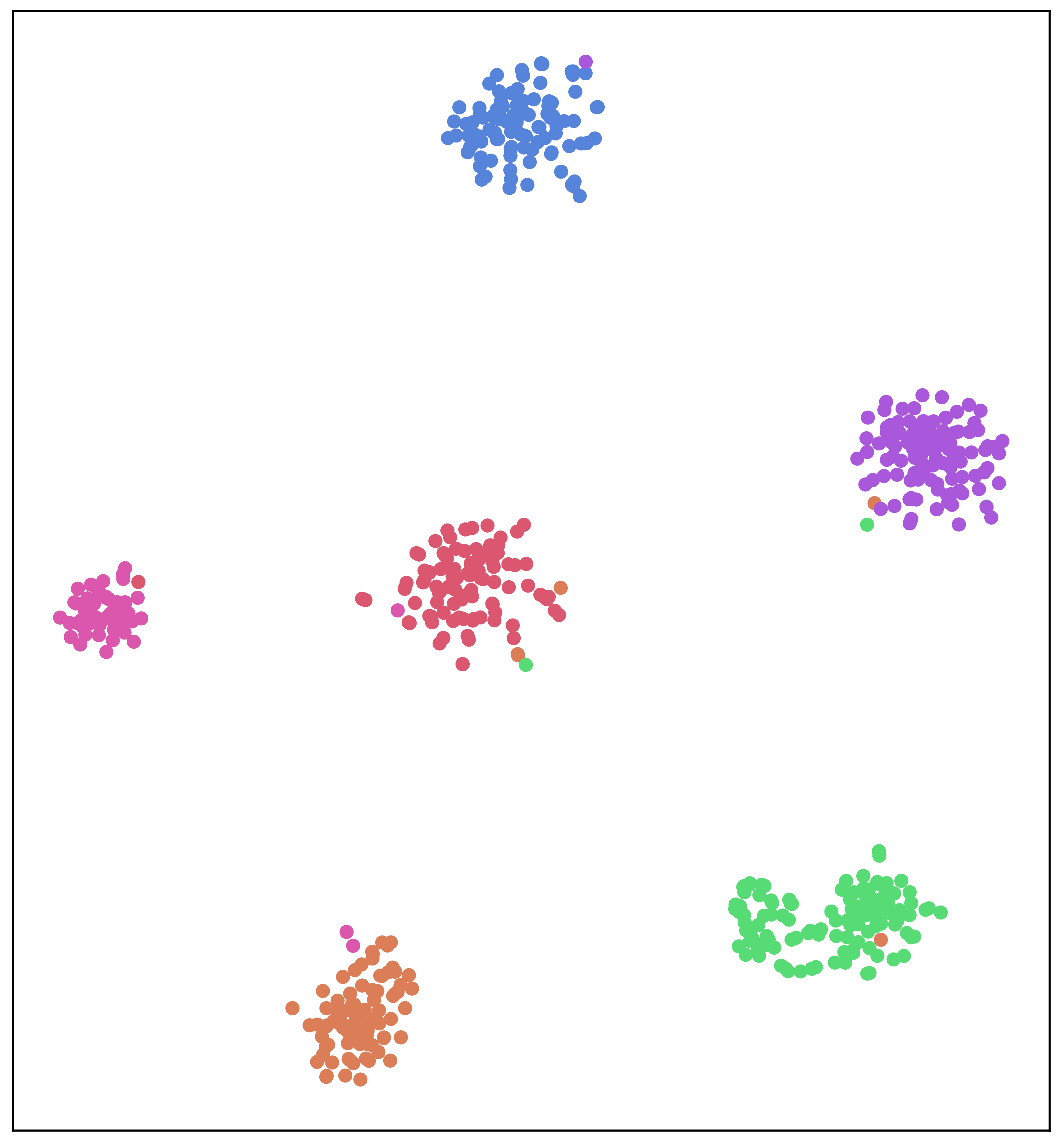}   
		\end{minipage}
	}
	\caption{t-SNE Visualization after 40 training epochs on target features for a randomly selected subset of 6 \textbf{DomainNet} (Rw$\rightarrow$Cl) Classes. Comparison of results with two baselines: Source Model Only and AaD~\cite{yang2022attracting}. Each color in the graphs represents a class of samples. It is evident from the visualizations that the proposed method surpasses both "Source model only" and "AaD" in terms of qualitative performance. This superiority is demonstrated through the generation of intrinsic local consistency and clear intra-class boundaries.} 
	\label{t-sne}  
\end{figure*}

In this section, we assess both the quality and quantity of pseudo-labels generated by each component of our method, comparing them with the source model alone and SHOT. Pseudo-label quantity is measured by the ratio of selected samples to the total samples, while pseudo-label quality is defined as the precision of the selected samples. The results are shown in Table~\ref{appendix_qq}. As seen, using the original source model generates good pseudo-label quality within the selected group, but only a small number of samples satisfy the high confidence condition. On the other hand, SHOT and PA select a large number of samples but with a relatively poor quality of approximately 80\%. In comparison, PA+HCPR achieves both good pseudo-label quality (90.76\%), and a substantial quantity of pseudo-labels (24.65\%). When comparing HCPR only and PA only, we observed that PA generates nearly four times as many pseudo-labels as HCPR but with lower quality. This suggests the presence of significant noise in the pseudo-labels generated by PA. 

The training progress to both the quantity and quality of pseudo-labels can be shown in Figure~\ref{appendix_course}. Our findings revealed that in the initial step with PA (0-9th epochs), there is a significant increase in the quantity of pseudo labels, albeit accompanied by a gradual decrease in their quality. However, with the assistance of HCPR (after 9th epch, before 10th epoch), the quality of pseudo-labels experiences a significant increase, accompanied by a substantial quantity. In the subsequent third step involving FM (10-40th epochs), the quality of pseudo labels has a gradual improvement, which subsequently stabilizes at a consistent level.

\subsection{Incorporating the Proposed Method into Existing Approaches}
\label{exp: SHOT+ours}

The proposed method can be seamlessly integrated into existing network architectures, such as SHOT~\cite{liang2020we} and AaD~\cite{yang2022attracting}. Specifically, we replace the pre-adaptation phase in our first step with SHOT and AaD, resulting in the combined approach referred to as ``SHOT+Ours" and ``AaD+Ours". The integration process can be summarized as follows: first, pseudo labels are generated using SHOT's unsupervised nearest class centroid approach and AaD's feature clustering and cluster assignment approach. Then, to refine these pseudo labels and address potential noise and inaccuracies, we utilize hypothesis consolidation of prediction rationale. The refined pseudo-label set is used as the labeled dataset, while the remaining samples are treated as unlabeled. Consequently, the SFUDA problem is transformed into a semi-supervised learning problem. 
The experimental results, as shown in Table~\ref{intergation_SHOT}, demonstrate the superiority of the proposed method integrated into the SHOT and AaD objectives. Across the Office-Home (Avg. $\uparrow$ 1.6\% and $\uparrow$ $0.7\%$), VisDA-C (Avg. $\uparrow$ 3.9\% and $\uparrow$ $0.4\%$), and DomainNet-126 (Avg. $\uparrow$ 2.8\% and $\uparrow$ $3.7\%$) datasets, the integrated approach consistently outperforms the baseline of SHOT and AaD. This indicates that our method complements existing SFUDA baselines and consistently improves their performance by incorporating our approach as a replacement for the model pre-adaptation phase.

\subsection{Visualization}
In t-SNE visualization, we compare the results with the state before adaptation by examining three approaches: source model only, AaD~\cite{yang2022attracting}, and our method shown in Figure~\ref{t-sne}. The source model only demonstrates shortcomings, experiencing false predictions within each class and struggling to establish clear intra-class boundaries. While AaD generally achieves accurate predictions within each class, it falls short in generating clear intra-class boundaries. In contrast, our method excels in achieving accurate predictions within each class and successfully generates distinct intra-class boundaries, which showcases its ability to enhance prediction accuracy and produce well-defined intra-class boundaries. 

\section{Limitation and Future Work}
The current approach relies on having access to the entire target training set to perform crucial steps like pre-adaptation and identifying the reliable pseudo-labeled set. However, in real-world applications, online adaptation is often more desirable as it doesn't require holding a large number of target examples. As part of our future work, we aim to extend the key idea of this research to the online streaming setting. By doing so, we can develop a methodology that adapts in real-time to incoming data, allowing for more efficient and effective adaptation in dynamic environments. This extension will enhance the applicability and practicality of the proposed approach in various domains.

\section{Conclusion}
In conclusion, this paper introduces a novel approach for Source-Free Unsupervised Domain Adaptation (SFUDA), where a model needs to adapt to a new domain without access to target domain labels or source domain data. By considering multiple prediction hypotheses and analyzing their rationales, the proposed method identifies the most likely correct hypotheses, which are then used as pseudo-labeled data for a semi-supervised learning procedure. The three-step adaptation process, including model pre-adaptation, hypothesis consolidation, and semi-supervised learning, ensures optimal performance. Experimental results demonstrate that the proposed approach achieves state-of-the-art performance in the SFUDA task and can be seamlessly integrated into existing methods to enhance their performance.

\section*{Acknowledgments}
We would like to thank the anonymous reviewers for their helpful comments.
This work is supported by the Centre for Augmented Reasoning.



 

\bibliography{main.bbl}
\bibliographystyle{IEEEtran}











\newpage

 





\end{document}